\documentclass{article}

\usepackage{arxiv}

\usepackage[utf8]{inputenc} % allow utf-8 input
\usepackage[T1]{fontenc}    % use 8-bit T1 fonts
\usepackage{hyperref}       % hyperlinks
\usepackage{url}            % simple URL typesetting
\usepackage{booktabs}       % professional-quality tables
\usepackage{amsmath,amssymb,amsfonts,bm}       % blackboard math symbols
\usepackage{url}      % Package for URL formatting
\usepackage[numbers,sort&compress]{natbib}

\usepackage{nicefrac}       % compact symbols for 1/2, etc.
\usepackage{microtype}      % microtypography
\usepackage{lipsum}
\usepackage{graphicx}
\usepackage{subcaption}
\graphicspath{ {./fig/} }
\usepackage[ruled,vlined,linesnumbered]{algorithm2e}

\usepackage{amsmath}
\usepackage{multirow}

\usepackage{cancel}

\usepackage[capitalize]{cleveref}

\usepackage[table,xcdraw]{xcolor}
\usepackage{tabularx}

\title{Temporally Unified Adversarial Perturbations for Time Series Forecasting}
\author{
  Ruixian Su \\
  School of Management \\
  Huazhong University of Science and Technology \\
  \texttt{ruixiansu@hust.edu.cn} \\
  \And
  Yukun Bao \\
  School of Management \\
  Huazhong University of Science and Technology \\
  \texttt{yukunbao@hust.edu.cn} \\
  \And
  Xinze Zhang\thanks{Corresponding author: Xinze Zhang.} \\
  School of Computer Science and Technology\\
  Huazhong University of Science and Technology \\
  \texttt{xinze@hust.edu.cn} \\
}

\newcommand\ie{\textit{i.e.}}
\newcommand\eg{\textit{e.g.}}

\newcommand\x{\bm{x}}
\newcommand\y{\bm{y}}

\newcommand\w{\bm{\omega}}
\newcommand\bc{\bm{c}}

\newcommand\m{\bm{m}}
\newcommand\bp{\bm{p}}

\newcommand\bv{\bm{v}}
\newcommand\bg{\bm{g}}

\newcommand\argmax{\operatorname{argmax}}

\newcommand\sign{\operatorname{sign}}

\newcommand\clip{\operatorname{clip}}

\newcommand\bdelta{\bm{\delta}}

\begin{document}
\maketitle

% 摘要

\begin{abstract}
While deep learning models have achieved remarkable success in time series forecasting, their vulnerability to adversarial examples remains a critical security concern.
However, existing attack methods in the forecasting field typically ignore the temporal consistency inherent in time series data, leading to divergent and contradictory perturbation values for the same timestamp across overlapping samples. 
This temporally inconsistent perturbations problem renders adversarial attacks impractical for real-world data manipulation. 
To address this, we introduce Temporally Unified Adversarial Perturbations (TUAPs), which enforce a temporal unification constraint to ensure identical perturbations for each timestamp across all overlapping samples. 
Moreover, we propose a novel Timestamp-wise Gradient Accumulation Method (TGAM) that provides a modular and efficient approach to effectively generate TUAPs by aggregating local gradient information from overlapping samples.
By integrating TGAM with momentum-based attack algorithms, we ensure strict temporal consistency while fully utilizing series-level gradient information to explore the adversarial perturbation space.
Comprehensive experiments on three benchmark datasets and four representative state-of-the-art models demonstrate that our proposed method significantly outperforms baselines in both white-box and black-box transfer attack scenarios under TUAP constraints.
Moreover, our method also exhibits superior transfer attack performance even without TUAP constraints, demonstrating its effectiveness and superiority in generating adversarial perturbations for time series forecasting models.
\end{abstract}


% 第一章：引言
\section{Introduction}

Deep-learning-based time series forecasting has been widely deployed due to its superior predictive performance, serving as a pivotal tool for operations and risk management across critical sectors, such as smart grids~\cite{aslam_survey_2021}, manufacturing~\cite{xu_novel_2025}, and finance~\cite{olorunnimbe_deep_2023}.
Despite these advances, recent studies have revealed that deep learning models are vulnerable to adversarial examples, \ie, inputs with carefully crafted perturbations that lead to incorrect predictions~\cite{szegedy_intriguing_2014}.
This particularly indicates the adversarial vulnerability in time series forecasting applications, where malicious perturbations could mislead forecasting systems into making erroneous decisions with catastrophic consequences.
Consequently, investigating the reliability and robustness of time series forecasting models under adversarial attacks has garnered considerable attention.

While adversarial attacks in vision and language domains have been extensively studied, attacks in the forecasting field remain limited.
Early attempts at adversarial attacks on time series forecasting directly adapted image-based attack methods, such as the Fast Gradient Sign Method (FGSM)~\cite{goodfellow_explaining_2015}, treating time series as one-dimensional images~\cite{mode_adversarial_2020,heinrich_targeted_2024}.
However, such adaptations neglect the unique temporal dependencies in time series~\cite{belkhouja_dynamic_2023}. 
More recent works have begun to tailor adversarial attack strategies specifically for time series forecasting, such as the importance measuring method AAIM~\cite{wu_small_2022}, the temporal similarity constraint method TCA~\cite{shen_temporal_2025}, and the attack direction selection method AAJM~\cite{jiao_gradient-based_2024}, which have significantly enhanced the effectiveness of adversarial attacks in the time series forecasting domain.

Despite their efficacy, existing methods overlook the auto-regressive sliding-window characteristic of time series forecasting, which makes forecasting tasks inherently distinct from other tasks. 
This mechanism implies that a historical value at a specific timestamp is repeatedly sampled and observed within multiple input samples.
Since conventional attack frameworks optimize perturbations for each sample in isolation, they inevitably generate divergent, often contradictory, perturbation values for the same timestamp across different samples, leading to the "temporally inconsistent perturbations" problem.
This lack of temporal consistency renders such adversarial examples impractical for real-world data manipulation, thereby hindering a realistic evaluation of the adversarial robustness of time series forecasting systems.
% , as illustrated in Figure~\ref{fig:Introduction}. 

% xinze: This figure needs to be revised.
% \begin{figure}
%     \centering
%     \includegraphics[width=1\linewidth]{fig/Introduction.png}
%     \caption{Illustration of the sliding-window forecasting mechanism and the conflict of multi-valued perturbations at the same timestamp.}
%     \label{fig:Introduction}
% \end{figure}

To address these limitations, we propose a novel and practical concept of adversarial perturbations, namely Temporally Unified Adversarial Perturbations (TUAPs), which significantly degrade the accuracy of state-of-the-art forecasting models while explicitly enforcing a temporal unification constraint that ensures the perturbation at each specific timestamp is shared across all overlapping samples.
Since TUAPs impose a more stringent constraint on the perturbation space, generating effective TUAPs is more challenging than crafting traditional perturbations.
To effectively generate TUAPs, we further propose a modular and efficient Timestamp-wise Gradient Accumulation Method (TGAM), by aggregating local gradient information from overlapping samples to optimize perturbations under the temporal consistency constraint.
Through integrating TGAM with momentum-based attack algorithms, we demonstrate that our proposed method can be easily incorporated into existing attack frameworks to enhance their effectiveness for time series forecasting.
Our main contributions are summarized as follows:
\begin{itemize}
    \item We pioneer the investigation of Temporally Unified Adversarial Perturbations (TUAPs) for time series forecasting, demonstrating that state-of-the-art forecasting models can be effectively attacked across all overlapping samples using temporally unified perturbations that maintain practical feasibility.
    \item We propose the Timestamp-wise Gradient Accumulation Method (TGAM) to effectively generate TUAPs, which efficiently aggregates gradient information from overlapping samples to optimize perturbations under the temporal consistency constraint.
    \item We conduct extensive experiments demonstrating that our proposed method significantly outperforms all baselines in both white-box and black-box attack scenarios under TUAP constraints, and even exceeds unconstrained baselines in transfer attack scenarios, demonstrating its effectiveness and superiority in attacking time series forecasting models.
\end{itemize}

The remainder of this paper is organized as follows.
Section 2 provides preliminaries on time series forecasting and reviews related work on adversarial attacks.
Section 3 details the motivation of this work, the definition of TUAPs, and the implementation of our proposed TGAM.
Section 4 presents the experimental setup, results, and analysis.
Finally, Section 5 concludes the paper and discusses future research.
% 第二章：相关工作

\section{Related Work}

\subsection{Preliminaries}

\begin{figure}[htbp]
    \centering
    \includegraphics[width=\linewidth]{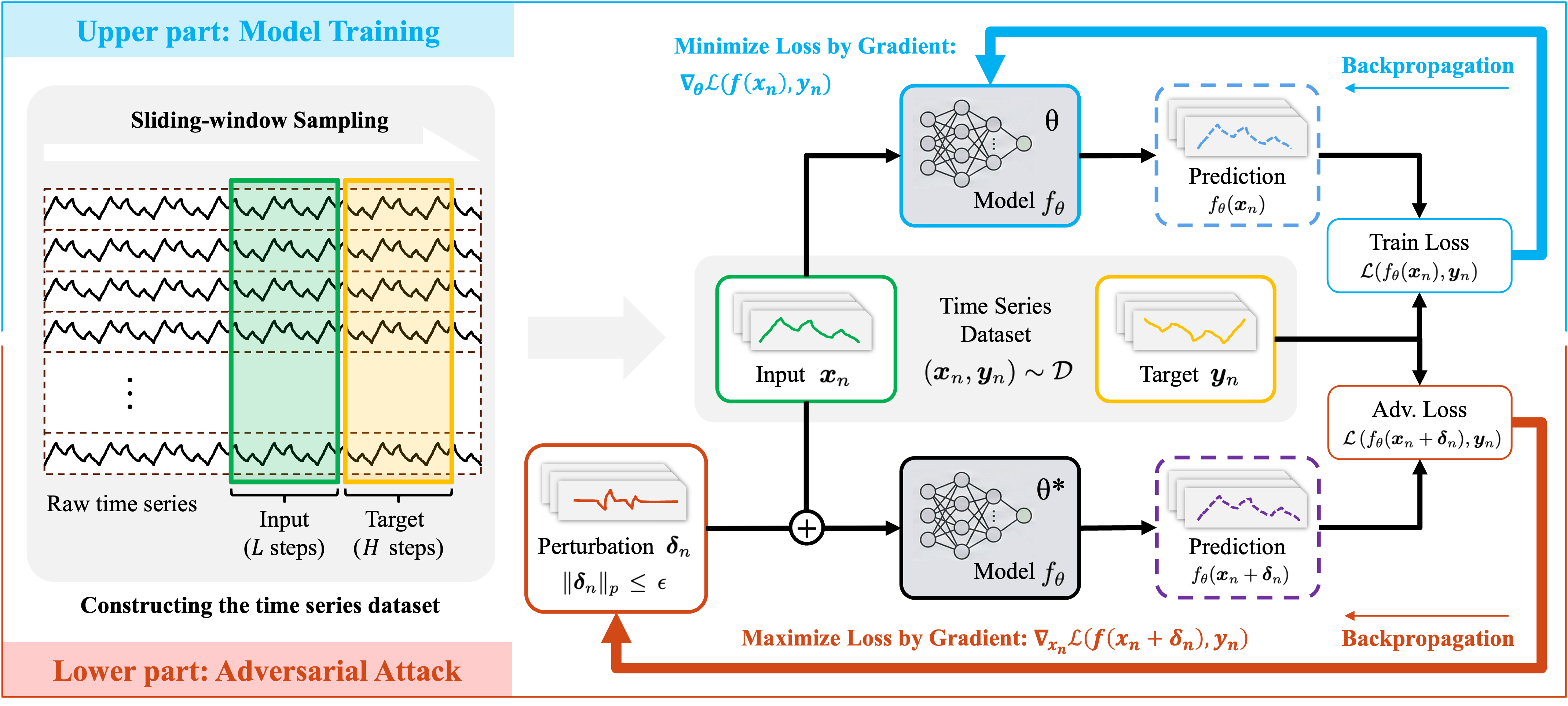}
    \caption{Illustration of the preliminaries of time series forecasting and adversarial attack on time series forecasting.}
    \label{fig:Preliminary}
\end{figure}

As shown in \cref{fig:Preliminary}, in a general time series forecasting task, the objective is to predict a sequence of future observations based on a given sequence of historical data.
Typically, consider a raw time series $\mathcal{V} = \{\bv_1, \bv_2, \dots, \bv_t, \dots, \bv_T\} \in \mathbb{R}^{T \times D}$, where $T$ is the total length of the time series, $\bv_t \in \mathbb{R}^D, \bv_t = [v_t^1, v_t^2, \ldots, v_t^D]$ denotes the observation vector at time $t$, and $D$ is the dimension (number of variables) of each observation.
Forecasting models employ a sliding-window mechanism to consecutively sample $\mathcal{V}$ without strides\footnote{We provide a dynamical visualization of what this non-stride sampling does in the link: \url{https://github.com/Simonnop/time_series_sampling/blob/main/README.md}} to construct a time series dataset $\mathcal{D}$ that consists of $N$ input-target pairs $\{(\x_n,\y_n)\}_{n=1}^N$.

Specifically, let $L$ denote the input lookback window length and $H$ denote the forecasting horizon.
The input $\x_n \in \mathbb{R}^{L \times D}$ and target $\y_n \in \mathbb{R}^{H \times D}$ are sampled as $\x_n = [\bv_n, \bv_{n+1}, \ldots, \bv_{n+L-1}]$ and $\y_n = [\bv_{n+L}, \bv_{n+L+1}, \ldots, \bv_{n+L+H-1}]$, respectively.
For brevity, let $\bv_{n:n+L-1} = [\bv_n, \bv_{n+1}, \ldots, \bv_{n+L-1}]$ denote the time series values from timestamp $n$ to timestamp $n+L-1$, so that $\x_n = \bv_{n:n+L-1}$, $\y_n = \bv_{n+L:n+L+H-1}$, and $N = T-L-H+1$.

Deep forecasting models learn a mapping function $f_\theta: \mathcal{X} \to \mathcal{Y}$, where $\mathcal{X} \subseteq \mathbb{R}^{L \times D}$ and $\mathcal{Y} \subseteq \mathbb{R}^{H \times D}$ are the input and target spaces, respectively.
The objective is to minimize a loss function $\mathcal{L}(\cdot,\cdot)$ (\eg, mean squared error~\cite{wu_timesnet_2022, liu_itransformer_2023}) between the prediction $\hat{\y}_n = f_\theta(\x_n)$ and the ground truth ${\y}_n$ with the optimal parameters $\theta^*$:
\begin{equation}
    \label{eq:training_objective}
\theta^* = \underset{\theta}{\operatorname{argmin}} \, \mathcal{L}(\hat{\y}_n, \y_n).
\end{equation}
In contrast, adversarial attacks aim to find an adversarial example ${\x}'_n = \x_n + \bdelta_n$ within a small neighborhood of the original input ${\x}_n$ that causes erroneous predictions, where $\bdelta_n$ is the adversarial perturbation.
For each sample $\x_n$, the objective is to maximize the loss function $\mathcal{L}(\cdot,\cdot)$ between the prediction $\hat{\y}^\prime_n = f_\theta({\x}^\prime_n)$ and the ground truth $\y_n$ with the optimal perturbation ${\bdelta}_n^*$:
\begin{equation}
    \label{eq:adversarial_objective}
    {\bdelta}_n^* = \underset{{\bdelta}_n}{\operatorname{argmax}} \, \mathcal{L}\left( \hat{\y}^\prime_n, \y_n \right) 
    \quad \text{s.t.} \quad \|{\bdelta}_n\|_p \leq \epsilon,
\end{equation}
where $\|\bdelta_n\|_p \leq \epsilon$ constrains the perturbation magnitude, \ie, the $L_p$ norm of the perturbation $\bdelta_n$ must be less than or equal to a threshold $\epsilon$, with $L_\infty$ being the most commonly adopted norm~\cite{dingBlackBoxAdversarialAttack2023, modeAdversarialExamplesDeep2020}.

\subsection{Foundational Adversarial Attack Methods}
Adversarial attacks serve as an important tool for assessing the adversarial robustness of deep learning models. 
Since \citet{szegedy_intriguing_2014} first identified the vulnerability of Deep Neural Networks (DNNs), research in this field has flourished, primarily within the computer vision domain. 
As a foundational work, \citet{goodfellow_explaining_2015} propose the Fast Gradient Sign Method (FGSM), which generates adversarial perturbations via a single-step update along the direction of the loss gradient:
\begin{equation}
\bdelta_n = \epsilon \cdot \sign(\nabla_{\x_n} \mathcal{L}(\hat{\y}_n, \y_n)) = 
\begin{cases} 
\epsilon, & \text{if } \nabla_{\x_n} \mathcal{L}(\hat{\y}_n, \y_n) > 0 \\
0, & \text{if } \nabla_{\x_n} \mathcal{L}(\hat{\y}_n, \y_n) = 0 \\
-\epsilon, & \text{if } \nabla_{\x_n} \mathcal{L}(\hat{\y}_n, \y_n) < 0 
\end{cases},
\label{eq:FGSM}
\end{equation}
where $\sign(\cdot)$ is the sign function, and $\nabla_{\x_n} \mathcal{L}(\hat{\y}_n, \y_n)$ is the gradient of the loss $\mathcal{L}(\hat{\y}_n, \y_n)$ with respect to $\x_n$. 
While efficient, this single-step attack may be limited in effectiveness on complex model architectures due to its simplistic optimization of adversarial perturbations.

To overcome the limitations of FGSM, \citet{kurakin_adversarial_2017} propose an iterative variant known as the Basic Iterative Method (BIM). 
Let $k \in \{0,1,\ldots,K-1\}$ denote the iteration index and $K$ denote the maximum number of iterations. 
$\x^{\prime,k}_n$ and $\bdelta^{k}_n$ denote the adversarial example and adversarial perturbation at the $k$-th iteration, BIM can be formulated as:
\begin{align}
\x'^{,k+1}_n & = \x_n + \bdelta_n^{k+1}, 
\\
\bdelta^{k+1}_n & = {\clip(\tilde{\bdelta}_n^{k+1}, \epsilon)}  = 
\begin{cases} 
\epsilon, & \text{if } \tilde{\bdelta}_n^{k+1} >   \epsilon,\\
-\epsilon, & \text{if } \tilde{\bdelta}_n^{k+1} <  - \epsilon,\\
\tilde{\bdelta}_n^{k+1}, & \text{otherwise},
\end{cases} \label{eq:Clip}
\\
\tilde{\bdelta}_n^{k+1} & =\bdelta_n^{k} + \alpha \cdot \sign(\nabla_{\x'^{,k}_n} \mathcal{L}(f_\theta(\x'^{,k}_n), \y_n)), \label{eq:BIM}
\end{align}
where $\x'^{,0}_n = \x_n$, $\bdelta^{0}_n = \bm{0}$, and $\alpha$ is the step size with $\alpha < \epsilon$..
This iterative approach allows for a more thorough exploration of the adversarial space. 
Facing the issue of getting trapped in local extrema during optimization, \citet{madry_towards_2019} propose Projected Gradient Descent (PGD), which extends BIM by introducing random initialization. 
By starting from a random point within the $\epsilon$-neighborhood, such as $\x'^{0}_n = \x_n + \operatorname{Uniform}(-\epsilon, \epsilon)$, PGD provides a stronger and more reliable attack than BIM. 

Furthermore, to improve the transferability of adversarial examples, \citet{dong_boosting_2018} propose Momentum Iterative FGSM (MI-FGSM) by integrating a momentum term into the iterative update of BIM, which stabilizes the update direction and helps escape poor local maxima:
\begin{align}
\tilde{\bdelta}_n^{k+1} & = \bdelta_n^{k} + \alpha \cdot \sign(\m^{k}_n),  \label{eq:mi}
\\
\m^{k+1}_n & = \mu \cdot \m^{k}_n + \frac{\nabla_{\x'^{,k}_n} \mathcal{L}(f_\theta(\x'^{,k}_n), \y_n)}{\|\nabla_{\x'^{,k}_n} \mathcal{L}(f_\theta(\x'^{,k}_n), \y_n)\|_1},
\end{align}
where $\m^{k}_n$ is the accumulated momentum term at $k$-th iteration, $\m^{0}_n = \bm{0}$, and $\mu$ is the decay factor.

Building on these classical methods, numerous variants have been proposed focusing on gradient refinement~\cite{linNesterovAcceleratedGradient2019, wangEnhancingTransferabilityAdversarial2021}, ensemble surrogate models~\cite{liuDelvingTransferableAdversarial2017}, and input transformation techniques~\cite{xieImprovingTransferabilityAdversarial2019, dongEvadingDefensesTransferable2019}, leading to enhanced attack effectiveness in white-box settings and improved adversarial transferability in black-box settings.
While these methods provide a solid technical foundation for adversarial attacks, they are primarily designed for image-based tasks and do not explicitly account for the unique characteristics of time series forecasting.

\subsection{Adversarial Attacks for Time Series Forecasting}

Compared to the extensive research on adversarial attacks in the image domain, studies focusing on time series forecasting tasks have emerged more recently and remain relatively limited.

Initially, the primary approach was to adapt well-established adversarial attack techniques directly from the image domain to time series models by replacing the input $\x_n$ (from images to time series) and the loss function $\mathcal{L}$ (from Cross-Entropy classification loss to Mean Squared Error regression loss):
\begin{equation}
    \label{eq:mseloss}
    \mathcal{L}\left( \hat{\y}^{\prime}_n, \y_n \right)  = \frac{1}{H} \sum_{h=1}^{H} (\hat{\y}^{\prime}_{n,h} - \y_{n,h})^2, 
\end{equation}
where $\y_{n,h} = \bv_{n+L+h-1}$ and $\hat{\y}^{\prime}_{n,h}$ are the ground-truth and prediction values at the $h$-th horizon step of $\y^{\prime}_n$, respectively. 
Following this approach, \citet{mode_adversarial_2020} transfer FGSM and BIM to craft adversarial multivariate time series examples for CNN, LSTM, and GRU, demonstrating that adversarial examples exhibit effectiveness and transferability in time series forecasting.
Similarly, \citet{heinrich_targeted_2024} utilize PGD to construct adversarial examples for CNN and LSTM in wind power forecasting tasks.
In addition to gradient-based methods, \citet{wang_investigation_2023} formulate the generation of adversarial perturbations as a discrete optimization problem and utilize Bayesian Optimization (BO) to explore perturbations on CNN, LSTM, and Attention-based TCN.
These investigations highlight the vulnerability of deep learning models in time series forecasting domains.

Beyond directly applying existing methods, researchers have made efforts to improve these classical methods to better suit the forecasting domain, primarily focusing on preserving the imperceptibility of adversarial attacks in time series.
A natural approach is to selectively perform adversarial attacks on different time steps.
\citet{wu_small_2022} propose the Adversarial Time Series Generator (ATSG), which is based on FGSM but replaces the MSE loss in \cref{eq:mseloss} with a Mean Absolute Error (MAE) function:
\begin{equation}
  \mathcal{L}_{\text{ATSG}}  = \frac{1}{H} \sum_{h=1}^{H} |\hat{\y}^{\prime}_{n,h} - \y_{n,h}|.
\end{equation}
The authors~\cite{wu_small_2022} further proposed the Adversarial Attack with Importance Measuring (AAIM), which selectively applies perturbations to critical time steps within each input sample.
AAIM first generates candidate perturbations $\bdelta_n$ with ATSG, then calculates importance scores for each time step using a masking operation similar to textual adversarial attacks~\cite{renGeneratingNaturalLanguage2019,zhangCraftingAdversarialExamples2021}, and finally retains perturbations only on the most important time steps while resetting others to their benign values, thereby improving attack imperceptibility.

Considering selective adversarial attacks on different input samples, \citet{jiao_gradient-based_2024} propose a sample selection method and an Attack Direction Judgment Method (ADJM).
Specifically, the authors train a binary Support Vector Machine (SVM) to perform sample selection, determining whether each sample should be perturbed.
Then, ADJM employs FGSM with the objective of maximizing or minimizing the forecasting value:
\begin{equation}
  \mathcal{L}_{\text{ADJM}} = {\lambda}_n  \cdot \hat{\y}^{\prime}_{n,h},
\end{equation}
where ${\lambda}_n \in \{1, -1\}$ denotes the attack direction and is predicted via another binary SVM for each selected sample $\x_n$.

Besides selective adversarial attacks, similarity between adversarial and benign samples has been considered to constrain perturbations.
\citet{shen_temporal_2025} propose Temporal Characteristics-based Attack (TCA) by modifying the numerical comparison in BIM's clip operation with cosine similarity comparison, replacing $\clip(\cdot, \epsilon)$ in \cref{eq:Clip} as:
\begin{equation}
\clip_{\text{TCA}}(\tilde{\bdelta}_n^k, \epsilon) = \begin{cases} 
    \epsilon, & \text{if } \operatorname{sim}(\x_n, \x_n + \epsilon) > \max(\operatorname{sim}(\x_n, \x_n - \epsilon),\operatorname{sim}(\x_n, \x_n + \tilde{\bdelta}_n^k)), \\
 -\epsilon, & \text{if } \operatorname{sim}(\x_n, \x_n - \epsilon) > \max(\operatorname{sim}(\x_n, \x_n + \epsilon),\operatorname{sim}(\x_n, \x_n + \tilde{\bdelta}_n^k)), \\
\tilde{\bdelta}_n^k, & \text{otherwise},
\end{cases}
\end{equation}
where $\operatorname{sim}(\cdot,\cdot)$ is the cosine similarity function. Through this operation, TCA effectively crafts perturbed samples $\x'_n$ that maintain higher similarity to the original samples $\x_n$ than previous attack methods.

Despite these efforts, existing adversarial attack methods for time series forecasting still follow the sample-independent attack paradigm derived from the image domain, generating perturbations independently for each input sample.
Although imperceptibility is improved by selectively attacking critical time steps within each sample or by constraining similarity between adversarial and benign samples, these works all focus on sample-level imperceptibility.
This is analogous to adversarial attacks on images that tolerate pixel-level noise invisible to human perception independently for each image, but it neglects the intrinsic temporal consistency requirement across consecutive samples in time series data.
Specifically, due to the sliding-window mechanism in time series forecasting, the same timestamp appears in multiple overlapping samples, yet existing methods may assign contradictory perturbation values to the same timestamp across different samples, making such attacks physically unrealizable in real-world data manipulation scenarios.
Therefore, this study addresses this critical limitation by introducing temporally unified adversarial perturbations that ensure consistency across all overlapping samples.

% 改成说：“已有的方法”
% \begin{figure}[!t]
%     \centering
%     \includegraphics[width=1\linewidth]{fig/consistent.png}
%     \caption{Comparison between independent sample-wise attacks (with conflicting perturbations at the same time point) and the proposed global consistent perturbation scheme.}
%     \label{fig:consistent}
% \end{figure}

% 第三章：方法论

\section{Methodology}

In this section, we detail the definition and generation of Temporally Unified Adversarial Perturbations (TUAPs).
We begin by discussing the core motivations behind TUAPs, specifically addressing the temporally inconsistent perturbations issue caused by existing adversarial attacks in the forecasting domain.
Next, we provide a formal definition of TUAPs, introducing a constraint for temporal consistency across overlapping samples.
Then, with the updated adversarial objective, we introduce the Timestamp-wise Gradient Sign Method (TGSM) and the Timestamp-wise Gradient Accumulation Method (TGAM) to effectively generate TUAPs.
Finally, we present MI-TGAM as our final implementation, which incorporates a momentum-based update mechanism to enhance attack effectiveness and transferability.

% \xinze{This part can be further supported by a toy experiment.}
\subsection{Motivation}
In the field of time series forecasting, adversarial attacks must account for temporal consistency across consecutive samples to remain practically imperceptible.
Under the sliding-window sampling mechanism, which is the fundamental principle for segmenting the raw time series to construct datasets, an observed value at a specific timestamp is shared across multiple overlapping samples.
However, existing adversarial attack methods generate different adversarial perturbations for the same timestamp across overlapping samples.
This leads to a crucial problem that we refer to as ``temporally inconsistent perturbations'', making it impossible to assign a unique perturbation to a specific timestamp in reality.

Specifically, consider two adjacent input samples $\x_n$ and $\x_{n+1}$ derived from the raw time series $\mathcal{V} = \{\bv_1, \bv_2, \dots, \bv_T\}$, where $\x_n = \bv_{n:n+L-1}$ and $\x_{n+1} = \bv_{n+1:n+L}$ share the overlapping segment $\bv_{n+1:n+L-1}$.
When generating adversarial examples with existing methods, since the loss function values for $\x_n$ and $\x_{n+1}$ are different as $\mathcal{L}(\hat{\y}_n, \y_n) \neq \mathcal{L}(\hat{\y}_{n+1}, \y_{n+1})$, leading to inconsistent gradients as $\nabla_{\x_n} \mathcal{L}(\hat{\y}_n, \y_n) \neq \nabla_{\x_{n+1}} \mathcal{L}(\hat{\y}_{n+1}, \y_{n+1})$. 
Even though some gradient components may be clipped to the same magnitude $\epsilon$, this still leads to an extremely high probability of producing divergent perturbations for the shared timestamps.

This violates an important and natural constraint in the time series forecasting domain: once a timestamp is observed, its value becomes part of the fixed historical record and cannot be changed retroactively. 
Since the observations $\bv_{n+1:n+L-1}$ are historical and shared between both samples, once perturbations are applied to these timestamps, they become part of the fixed history and must be identical in both samples to maintain temporal consistency.
In contradiction to this requirement, existing adversarial attack methods generate inconsistent perturbations for the same timestamp, resulting in $\bdelta_{n,l+1} \neq \bdelta_{n+1,l}$, 
where $\bdelta_{n,l}$ denotes the generated adversarial perturbation for sample $\x_n$ at the $l$-th time step.
This temporally inconsistent perturbations problem poses several practical challenges when imposing adversarial perturbations on the input time series, such as the need for repeated perturbation injection for the same timestamp (which can be easily detected), and the same-timestamp adversarial perturbation selection problem (which remains unexplored).

To address the temporally inconsistent perturbations problem, we formally define Temporally Unified Adversarial Perturbations (TUAPs), which extend the adversarial attack scope from individual time series samples to the entire time series. 
We then introduce the Timestamp-wise Gradient Sign Method (TGSM), which provides a straightforward approach for generating TUAPs.
Additionally, to further enhance computational efficiency and attack effectiveness, we develop the Timestamp-wise Gradient Accumulation Method (TGAM), which provides a modular framework to improve attack performance.
Finally, we present MI-TGAM as our final implementation, which embeds the momentum-based update mechanism of MI-FGSM into TGAM to further enhance attack effectiveness and transferability.

\newtheorem{definition}{Definition}
\subsection{Temporally Unified Adversarial Perturbation}
To better highlight the difference between our proposed temporally unified adversarial perturbations and existing adversarial perturbations, we first formally summarize previous works and provide their conception of time series forecasting adversarial perturbations as follows.

\begin{definition}[Adversarial Perturbation]
    \label{def:adversarial_perturbation}
Given a subset of (test) time series samples $\mathcal{S} \in \mathcal{D}$ and a small constant $\epsilon > 0$, an adversarial perturbation is an additive noise in the adversarial perturbation space
\begin{equation}
    \mathcal{A} = \{ \bdelta \mid \exists \, (\x, \y) \in \mathcal{S}, \|\bdelta\|_\infty \leq \epsilon  \wedge \ell(f_\theta(\x), \y) \leq \xi 
    \wedge \ell(f_\theta(\x + \bdelta), \y) > \gamma  \},
\end{equation}
which consists of all possible perturbations $\bdelta$ that fool the forecasting model $f_\theta(\cdot)$ on $\x + \bdelta$, where $\ell(\cdot,\cdot)$ is a metric for evaluating forecasting error, and $\xi \leq \gamma$ are thresholds for acceptable and unacceptable forecasting quality, respectively.
\end{definition}

The constraint $\ell(f_\theta(\x), \y) \leq \xi \wedge \ell(f_\theta(\x + \bdelta), \y) > \gamma$ ensures that the forecasting model is reliable on clean data $\x$ but vulnerable to the adversarial example $\x + \bdelta$. 
In practice, $\ell(f_\theta(\x), \y) \leq \xi$ and $\ell(f_\theta(\x + \bdelta), \y) > \gamma$ are achieved by performing the optimization as shown in \cref{eq:training_objective} and \cref{eq:adversarial_objective}, respectively.

While \cref{def:adversarial_perturbation} is reasonable for individual samples in conventional settings, where an adversary typically seeks an optimal perturbation $\bdelta_n$ to maximize the forecasting error for each sample $\x_n$ independently, we argue that generating adversarial perturbations for multiple samples without considering temporal unification is inappropriate in time series forecasting, because the same timestamp shared by multiple samples should be perturbed identically to maintain temporal coherence.

Building upon the motivation discussed above, we formally define temporally unified adversarial perturbations for time series forecasting, which ensure that the perturbation applied to the same timestamp is shared across all overlapping samples.

\begin{definition}[Temporally Unified Adversarial Perturbations]
    \label{def:tuap}
    Within \cref{def:adversarial_perturbation}, a temporally unified adversarial perturbation is an additive noise in the adversarial perturbation space
\begin{equation}
     \mathcal{U} = \{ \bdelta_n \in \mathcal{A} \mid \forall\, \bv_t = \x_{n, l} = \x_{i,j} , \, \bdelta_{n, l} = \bdelta_{i,j} = \bp_t \wedge \forall\, p^d_t \in \bp_t, |p^d_t| \leq \epsilon \cdot |v_t^d| \},
\end{equation} 
where $\bv_t = \x_{n, l} = \x_{i,j}$ denotes that input samples $\x_n$ and $\x_{i}$ overlap at the historical value $\bv_t$ at timestamp $t$ (i.e., $\x_{n,l} = \x_{i,j} = \bv_t$). $\bp_t = [p_t^1, p_t^2, \ldots, p_t^D] \in \mathbb{R}^D$ denotes the perturbation applied to timestamp $t$ and remains identical across all overlapping samples ($\bdelta_{n,l} = \bdelta_{i, j} = \bp_t$), where $p_t^d$ is the perturbation for the $d$-th variable $v_t^d$ at timestamp $t$ with $|p^d_t| \leq \epsilon \cdot |v_t^d|$.
\end{definition}

Under the commonly applied non-stride sliding-window sampling mechanism for constructing the time series forecasting dataset $\mathcal{D}$, this definition is equivalent to imposing two additional constraints on \cref{def:adversarial_perturbation}.
The first constraint ensures {temporal consistency} that, for any adjacent samples $\x_n, \x_{n+1} \in \mathcal{S}$, we require $\bdelta_{n, l+1} = \bdelta_{n+1, l} = \bp_{n+l}$ to ensure that the same timestamp receives shared perturbations across overlapping samples.
The second constraint ensures {magnitude imperceptibility} that, the perturbation $\bp_t$ for each timestamp $t$ is variable-wisely bounded by a percentage $\epsilon$ of the corresponding input value $\bv_t$ as $|p_t^d| \leq \epsilon \cdot |v_t^d|$ for all $v_t^d \in \bv_t$.

\subsection{Adversarial Perturbation Generation}
Let $\tau$ denote the starting timestamp of the test dataset $\mathcal{S} = \{(\x_n, \y_n)\}_{n = \tau}^N$. 
Given a timestamp $t$ covered in $\mathcal{S}$, i.e., $\tau \leq t \leq N +L - 1$, let ${\mathcal{T}}_t = \{ (\x_i, \y_i) \in \mathcal{S} \mid \bv_t \in \x_i \}$ denote the set of overlapping samples whose inputs contain the observed value $\bv_t$, such as ${\mathcal{T}}_\tau = \{(\x_\tau, \y_\tau)\}$ and ${\mathcal{T}}_{\tau+1} = \{(\x_i, \y_i)\}_{i=\tau}^{\tau+1}$.    
The adversarial objective under the TUAP constraints at timestamp $t$ can be defined as:
\begin{equation}
    \begin{split}
    \bdelta_n^* &=
    [\bp_{n}^*, \bp_{n+1}^*, \ldots, \bp_{t}^*, \ldots, \bp_{n+L-1}^*], \\ 
    \bp^*_t &= 
    \underset{{\bp}_t}{\argmax} \sum_{\x_i \in \mathcal{T}_t} \mathcal{L}( f_\theta(\x_i + \bdelta_i), \y_i),
    \quad \text{s.t.} \quad \forall\, p^d_t \in \bp_t, |p^d_t| \leq \epsilon \cdot |v_t^d|.
    \end{split}
    \label{eq:global_objective}
\end{equation}
Notably, optimizing $\bp_t$ for sample $\x_n$ in \cref{eq:global_objective} requires the adversarial perturbations $\bdelta_i$ of other overlapping samples $\x_i \in \mathcal{T}_t$, making it challenging to ensure the TUAP property while simultaneously perturbing multiple samples with cascading effects.

To address this, it is intuitive and effective to calculate perturbations per timestamp rather than per sample, since a perturbation $\bp_t$ at a specific timestamp $t$ is unique and simultaneously affects all samples whose time range covers $t$.
Thus, we develop the Timestamp-wise Gradient Sign Method (TGSM), which provides a straightforward approach by optimizing each timestamp independently without considering perturbations at other timestamps.

Specifically, TGSM solves the optimization problem in \cref{eq:global_objective} by performing a single-step gradient ascent on the loss function with respect to the perturbation $\bp_t$ at timestamp $t$.
Let $\check{\bdelta}_{i,t}$ denote a zero vector with the same shape as $\x_i$ except that the value at timestamp $t$ is replaced by $\bp_t$.
Then, the optimal perturbation $\bp_t$ can be calculated as:
\begin{equation} 
    \begin{split}
    \bp_t &= \epsilon \cdot \operatorname{sign} (\bg_t) \cdot \bv_t, \\
    \bg_t &= \nabla_{\bv_t} \sum_{\x_i \in \mathcal{T}_t} \mathcal{L}( f_\theta(\x_i + \check{\bdelta}_{i,t}), \y_i ).
    \end{split}
    \label{eq:solve_t}
\end{equation}
By calculating $\bp_t$ for each timestamp $t \in \{n, n+1, \ldots, n+L-1\}$, the adversarial example can be generated as $\x_n^\prime = \x_n + \bp_{n:n+L-1}$. 
Through sequentially calculating and caching the perturbation $\bp_t$ at historical timestamps, for subsequent input samples such as $\x_{n+1}$, TGSM only needs to calculate the perturbation $\bp_{n+L}$ for the new observation $\bv_{n+L}$.

However, TGSM has several limitations. 
First, TGSM does not share the same computational advantage as traditional FGSM.
Let $\mathcal{O}(1)$ denote a single forward and backward propagation for calculating the gradient on the input sample $\x_i$, the time complexity of TGSM for generating all $\bp_{\tau:N+L-1}$ in the test set $\mathcal{S}$ is $\mathcal{O}(L\cdot|\mathcal{S}|)$, leading to an average time complexity of $\mathcal{O}(L)$ per sample, whereas FGSM has $\mathcal{O}(1)$ complexity per sample. 
Second, this single-step approach ignores the non-convex curvature of the DNN's loss landscape. 
More importantly, it always uses the gradient direction at the original input $\x$ and ignores the perturbations at historical timestamps, which may lead to suboptimal attack performance. 

To address these issues, we propose the Timestamp-wise Gradient Accumulation Method (TGAM), which performs multi-step iterative optimization with gradient accumulation.
Let $\bdelta_i^k$ denote the perturbation for sample $\x_i$ at iteration $k$, which is sliced from the timestamp-level perturbations as $\bdelta_i^k = \bp^k_{i:i+L-1}$.
$\bv_t^{\prime,k}$ denotes the adversarial example of $\bv_t$ at iteration $k$, constructed as $\bv_t^{\prime,k} = \bv_t + \bp^k_t$.
We sequentially calculate the gradient $\bg_t^k$ for all timestamps $t$ at iteration $k$ as:
\begin{equation}
    \bg_t^k = \sum_{\x_i \in \mathcal{T}_t} \nabla_{\bv_t^{\prime,k}} \mathcal{L}( f_\theta(\x_i^{\prime,k}), \y_n ),
    \label{eq:global_gradient}  
\end{equation}
where $\x_i^{\prime,k} = \x_i + \bdelta_i^k = \x_i + \bp^k_{i:i+L-1}$.

The proposed TGAM offers several benefits. 
First, by performing multi-step optimization, TGAM can better navigate the non-convex loss landscape of DNNs, leading to stronger adversarial perturbations.
Second, TGAM is simple, modular, and can be easily integrated with various momentum-based optimization techniques to further improve attack performance and transferability. 
Third, the gradient accumulation mechanism allows TGAM to leverage the collective gradient information from all overlapping samples, which shares similar insights with input transformation-based attacks, thereby enhancing attack effectiveness by finding a more optimal perturbation direction that considers the interactions between samples.

Moreover, TGAM is computationally efficient. 
Unlike existing input transformation-based attacks that require generating new transformed samples for each input, TGAM cleverly leverages the inherent structure of time series forecasting by using overlapping input samples as natural transformations for the current sample.
This avoids the additional overhead incurred by introducing extra samples in input transformation approaches.
Consequently, TGAM achieves an overall time complexity of $\mathcal{O}(|\mathcal{S}|)$ for generating all perturbations $\bp^k_{\tau:N+L-1}$ in $\mathcal{S}$, demonstrating computational efficiency comparable to traditional methods such as PGD.

\subsection{Implementation of Attacking Algorithm}
As discussed above, TGAM provides a modular framework to improve attack performance for time series forecasting.
To illustrate this, we present MI-TGAM, which incorporates the momentum-based update mechanism of MI-FGSM~\cite{dong_boosting_2018} into TGAM.
MI-FGSM has been a standard baseline attack algorithm for follow-up works, making it a natural choice for demonstrating the effectiveness of our approach.

In contrast to traditional MI-FGSM, which computes the gradient for each sample independently, MI-TGAM sequentially computes the gradient for each timestamp by aggregating the gradients from all overlapping samples. 
For each iteration $k$, we first update the adversarial example $\x_n^{\prime,k} = \x_n + \bp^k_{n:n+L-1}$ for each input sample $\x_n \in \mathcal{S}$ by concatenating the current timestamp-level perturbations.
Then, we calculate the sample-level gradient $\bc_{n}^k \in \mathbb{R}^{L \times D}$ for each sample $\x_n^{\prime,k}$ via backpropagation as:
\begin{equation}
    \bc_{n}^k = \nabla_{\x_n^{\prime,k}} \mathcal{L}\big(f_\theta(\x_n^{\prime,k}), \y_n\big),
\end{equation}
collecting the gradient cache $\mathcal{C}^k = \{\bc_n^k\}_{n=\tau}^N$ for the attacking set $\mathcal{S}$. 

Next, we perform timestamp-wise accumulation of the sample-level gradients to obtain the global gradient $\bg_t^k$ for each timestamp $t$ as in \cref{eq:global_gradient}.
For brevity, we illustrate this gradient accumulation process with a commonly used practice that the samples are generated without strides.
Specifically, let $\bc_{n,l}^k$ denote the $l$-th gradient vector in $\bc_n^k = [\bc_{n,1}^k, \bc_{n,2}^k, \ldots, \bc_{n,L}^k]$ corresponding to the $l$-th time step of the input sample $\x_n^{\prime,k}$, which is associated with timestamp $t = n + l - 1$. 
Taking into account the boundary conditions of the test set, the global gradient for timestamp $t$ is then computed by accumulating all sample-level gradients at that timestamp as:
\begin{equation}
\bg_{t}^k  = \begin{cases} 
    \sum_{l=1}^{t-\tau+1} \bc_{t-l+1, l}^k & \text{when}\quad \tau +L -1 > t \geq \tau, \\
    \sum_{l=1}^{L} \bc_{t-l+1, l}^k & \text{when}\quad  N \geq t \geq \tau+L-1, \\
    \sum_{j=t-L+1}^{N} \bc_{j, t-j+1}^k & \text{when}\quad N+L-1 \geq t > N.
\end{cases}
\end{equation}

Through this process, we construct the temporally unified gradient $\bg_{\tau:N+L-1}^k$ for the entire attack set.

Finally, we update the timestamp-level perturbation $\bp^k_t$ by performing a momentum-based update with the accumulated gradient $\bc_{n}^k$ for each  timestamp $t$, as follows:
\begin{equation}
\begin{split}
    \bp_t^{k+1} & = \clip\left(\bp_t^{k} + \alpha \cdot \sign(\w^{k}_t) \cdot \bv_t, \epsilon \cdot \bv_t\right),  \label{eq:mi-tgam}
\\
\w^{k+1}_t & = \mu \cdot \w^{k}_t + \frac{\bg_t^k}{\|\bg_t^k\|_1},
\end{split}
\end{equation}
where $\w^k_t$ is the momentum term for timestamp $t$ at iteration $k$.
After $K$ iterations, the optimized perturbation series $\bp_{\tau:N+L-1} \leftarrow \bp^K_{\tau:N+L-1}$ is sliced back into sample-level perturbations $\bdelta_n = \bp_{n:n+L-1}$ for each sample $\x_n \in \mathcal{S}$, forming the final TUAPs.

Through this implementation, MI-TGAM effectively exploits gradient information from multi-step optimization while maintaining the temporally unified constraint, resolving the temporally inconsistent perturbations problem while enhancing attack effectiveness for time series forecasting.

% xinze: this algorithm and figure need be corrected and polished.
% \input{algo/Algorithm 1}
% \begin{figure}[t]
%     \centering
%     \includegraphics[width=1\linewidth]{fig/GGAA-procedure.png}
%     \caption{Overall procedure of the proposed Global Gradient Accumulation Attack.}
%     \label{fig:gga}
% \end{figure}

% The overall procedure of the proposed MI-TGAM is summarized in Algorithm \ref{alg:gga}.
% 第四章：实验

\section{Experiment}

\subsection{Experimental Setup}

\paragraph{Datasets.} 
To evaluate the attack performance in different application secenarios under the definition of temporally unified adversarial perturbations, we conduct experiments on three commonly used real-world time series datasets~\cite{yi_frequency-domain_2023,lin_segrnn_2023,wu_timesnet_2022,liu_itransformer_2023}: 
ETT\footnote{\url{https://github.com/zhouhaoyi/ETDataset}}(electricity transformers dataset ETTh1), Electricity\footnote{\url{https://archive.ics.uci.edu/dataset/321/electricityloaddiagrams20112014}}  (electricity consumption dataset), and Traffic\footnote{\url{https://pems.dot.ca.gov/}}  (road occupancy dataset). 
For each dataset, we use 70\% of the data for training, 10\% for validation, and 20\% for testing, following the standard split protocol in time series forecasting literature.
The test set is used as the attack set for evaluating attack performance.

\paragraph{Models.} 
To validate the effectiveness and generalization of the attacks across different neural architectures, we select four state-of-the-art time series forecasting models with representative structures: the MLP-based FreTS~\cite{yi_frequency-domain_2023}, the RNN-based SegRNN~\cite{lin_segrnn_2023}, the CNN-based TimesNet~\cite{wu_timesnet_2022}, and the Transformer-based iTransformer~\cite{liu_itransformer_2023}.

\paragraph{Baselines.} 
We compare the proposed method (MI-TGAM) with several representative methods: FGSM \cite{goodfellow_explaining_2015}, BIM \cite{kurakin_adversarial_2017}, PGD \cite{madry_towards_2019}, MI-FGSM \cite{dong_boosting_2018}, ATSG \cite{wu_small_2022},
 ADJM \cite{jiao_gradient-based_2024}, TCA \cite{shen_temporal_2025}, and BO \cite{wang_investigation_2023}. 
Among these baselines, FGSM, BIM, PGD, and MI-FGSM are general adversarial attack methods widely used in computer vision, while ATSG, ADJM, TCA, and BO are specifically designed for time series forecasting tasks.
Notably, all these methods optimize perturbations for individual samples independently, which inherently leads to temporally inconsistent perturbations where the same timestamp receives contradictory values across different samples. 
For fair comparison, we adapt these methods to satisfy the TUAP constraint by randomly selecting one perturbation from the overlapping samples for each shared timestamp.

\paragraph{Attack Settings.} 
% 攻击参数
Following prior works \cite{wu_small_2022,shen_temporal_2025},
we use the $L_\infty$ norm to constrain the maximum perturbation magnitude.
For each variable in the input time series, we set the perturbation budget $\epsilon = 0.1$, which limits the perturbation to 10\% of the original magnitude for each variable: $|p_t^d| \leq \epsilon \cdot |v_t^d|$.
For iterative attack methods (including ours), we set the number of iterations $K=10$ and the step size $\alpha = \epsilon / K = 0.01$.
For momentum-based methods, we set the decay factor $\mu = 1.0$ following MI-FGSM~\cite{dong_boosting_2018}.
For other hyperparameters specific to each baseline method, we follow the settings in their original papers.
For the forecasting tasks, we set both the input length $L$ and the prediction horizon $H$ to 96, which is a standard setting in time series forecasting literature.

We adopt a cross-model transfer attack evaluation protocol, where each of the four models is used in turn as a surrogate model to generate adversarial examples that are then used to attack the other three target models.
When the surrogate model and target model are the same, the setting corresponds to a white-box attack; when they differ, the setting corresponds to a black-box transfer attack.
The attack performance is evaluated using two metrics: Mean Squared Error (MSE) and Mean Absolute Error (MAE).
To more clearly compare the attack effectiveness, we also report the degradation percentage, calculated as:
\begin{equation}
\text{Degradation}(\%) = \frac{\text{Error}_{\text{Attack}} - \text{Error}_{\text{Clean}}}{\text{Error}_{\text{Clean}}} \times 100\%,
\end{equation}
where $\text{Error}_{\text{Attack}}$ is the error on adversarial examples and $\text{Error}_{\text{Clean}}$ is the error on clean data.
A higher degradation percentage indicates a stronger attack.

% , ensuring that the perturbation at each timestamp remains identical across all overlapping samples.

% To this end, we evaluated three alignment strategies: ``First'' (using the perturbation from the earliest window), ``Last'' (using the perturbation from the latest window), and ``Random'' (randomly selecting from overlapping windows). 
% Our experiments reveal that the ``First'' and ``Last'' strategies exhibit high instability, frequently resulting in extreme cases of exceptionally high or poor attack performance due to their reliance on localized edge gradients. 
% In contrast, the ``Random'' strategy proves to be the most robust, consistently representing the average attack potency of traditional methods while avoiding extreme performance fluctuations. 
% In the experimental section, we primarily present results obtained from the ``Random'' method, while results for ``First'' and ``Last'' are provided in the Appendix.

\begin{table}[!t]
\centering
\caption{White-box attack results of for MI-TGAM and baseline methods across four models and three datasets. Best results are highlighted in \textbf{bold}.}
\label{tab:white-box}
\scriptsize
\resizebox{\textwidth}{!}{
    % 请在导言区添加：\usepackage{booktabs} 和 \usepackage{multirow}
\newcolumntype{L}{>{\raggedright\arraybackslash}X} % 左对齐
\newcolumntype{C}{>{\centering\arraybackslash}X} % 居中对齐
\newcolumntype{R}{>{\raggedleft\arraybackslash}X} % 右对齐

\begin{tabularx}{1.5\textwidth}{llCCCCCCCCCC}
\toprule
\multirow{2}{*}{Dataset} & \multirow{2}{*}{Method} & \multicolumn{5}{c}{MSE (with Degradation Percentage)} & \multicolumn{5}{c}{MAE (with Degradation Percentage)} \\ \cmidrule(lr){3-7} \cmidrule(lr){8-12}
 &  & FreTS & SegRNN & TimesNet & iTransformer & Average & FreTS & SegRNN & TimesNet & iTransformer & Average \\ \midrule

\multirow{10}{*}{ETT} & Clean & 0.411 (-0.0\%) & 0.377 (-0.0\%) & 0.446 (-0.0\%) & 0.398 (-0.0\%) & 0.408 (-0.0\%) & 0.420 (-0.0\%) & 0.400 (-0.0\%) & 0.452 (-0.0\%) & 0.412 (-0.0\%) & 0.421 (-0.0\%) \\
 & FGSM & 0.440 ($\uparrow$7.1\%) & 0.401 ($\uparrow$6.6\%) & 0.474 ($\uparrow$6.2\%) & 0.417 ($\uparrow$4.6\%) & 0.433 ($\uparrow$6.1\%) & 0.443 ($\uparrow$5.5\%) & 0.419 ($\uparrow$4.9\%) & 0.463 ($\uparrow$2.4\%) & 0.424 ($\uparrow$2.9\%) & 0.437 ($\uparrow$3.9\%) \\
 & BIM & 0.440 ($\uparrow$7.1\%) & 0.401 ($\uparrow$6.5\%) & 0.478 ($\uparrow$7.0\%) & 0.416 ($\uparrow$4.5\%) & 0.434 ($\uparrow$6.3\%) & 0.443 ($\uparrow$5.5\%) & 0.419 ($\uparrow$4.9\%) & 0.464 ($\uparrow$2.6\%) & 0.423 ($\uparrow$2.7\%) & 0.437 ($\uparrow$3.9\%) \\
 & PGD & 0.432 ($\uparrow$5.1\%) & 0.394 ($\uparrow$4.7\%) & 0.469 ($\uparrow$5.0\%) & 0.412 ($\uparrow$3.4\%) & 0.427 ($\uparrow$4.6\%) & 0.436 ($\uparrow$3.9\%) & 0.414 ($\uparrow$3.5\%) & 0.460 ($\uparrow$1.8\%) & 0.421 ($\uparrow$2.1\%) & 0.433 ($\uparrow$2.8\%) \\
 & MI-FGSM & 0.440 ($\uparrow$7.1\%) & 0.402 ($\uparrow$6.6\%) & 0.478 ($\uparrow$7.2\%) & 0.417 ($\uparrow$4.7\%) & 0.434 ($\uparrow$6.4\%) & 0.443 ($\uparrow$5.5\%) & 0.420 ($\uparrow$5.0\%) & 0.464 ($\uparrow$2.7\%) & 0.424 ($\uparrow$2.9\%) & 0.438 ($\uparrow$4.0\%) \\
 & ATSG & 0.438 ($\uparrow$6.6\%) & 0.400 ($\uparrow$6.3\%) & 0.472 ($\uparrow$5.8\%) & 0.415 ($\uparrow$4.1\%) & 0.431 ($\uparrow$5.7\%) & 0.443 ($\uparrow$5.6\%) & 0.421 ($\uparrow$5.3\%) & 0.464 ($\uparrow$2.6\%) & 0.425 ($\uparrow$3.1\%) & 0.438 ($\uparrow$4.1\%) \\
 & ADJM & 0.411 ($\uparrow$0.1\%) & 0.382 ($\uparrow$1.4\%) & 0.451 ($\uparrow$1.0\%) & 0.401 ($\uparrow$0.7\%) & 0.411 ($\uparrow$0.8\%) & 0.424 ($\uparrow$1.1\%) & 0.407 ($\uparrow$1.7\%) & 0.454 ($\uparrow$0.4\%) & 0.415 ($\uparrow$0.7\%) & 0.425 ($\uparrow$1.0\%) \\
 & TCA & 0.436 ($\uparrow$6.2\%) & 0.397 ($\uparrow$5.6\%) & 0.474 ($\uparrow$6.3\%) & 0.414 ($\uparrow$3.8\%) & 0.430 ($\uparrow$5.5\%) & 0.442 ($\uparrow$5.2\%) & 0.419 ($\uparrow$4.7\%) & 0.464 ($\uparrow$2.6\%) & 0.424 ($\uparrow$2.9\%) & 0.437 ($\uparrow$3.8\%) \\
 & BO & 0.415 ($\uparrow$1.0\%) & 0.377 ($\uparrow$0.1\%) & 0.450 ($\uparrow$0.8\%) & 0.405 ($\uparrow$1.6\%) & 0.412 ($\uparrow$0.9\%) & 0.427 ($\uparrow$1.6\%) & 0.401 ($\uparrow$0.3\%) & 0.457 ($\uparrow$1.1\%) & 0.420 ($\uparrow$1.8\%) & 0.426 ($\uparrow$1.2\%) \\
 & MI-TGAM & \textbf{0.454 ($\uparrow$10.5\%)} & \textbf{0.416 ($\uparrow$10.6\%)} & \textbf{0.508 ($\uparrow$13.8\%)} & \textbf{0.431 ($\uparrow$8.1\%)} & \textbf{0.452 ($\uparrow$10.9\%)} & \textbf{0.454 ($\uparrow$8.2\%)} & \textbf{0.434 ($\uparrow$8.4\%)} & \textbf{0.475 ($\uparrow$5.1\%)} & \textbf{0.434 ($\uparrow$5.3\%)} & \textbf{0.449 ($\uparrow$6.7\%)} \\
 \midrule
 
\multirow{10}{*}{ECL} & Clean & 0.195 (-0.0\%) & 0.206 (-0.0\%) & 0.259 (-0.0\%) & 0.154 (-0.0\%) & 0.204 (-0.0\%) & 0.283 (-0.0\%) & 0.291 (-0.0\%) & 0.341 (-0.0\%) & 0.245 (-0.0\%) & 0.290 (-0.0\%) \\
 & FGSM & 0.212 ($\uparrow$8.8\%) & 0.223 ($\uparrow$8.3\%) & 0.276 ($\uparrow$6.6\%) & 0.168 ($\uparrow$8.7\%) & 0.220 ($\uparrow$8.0\%) & 0.305 ($\uparrow$7.6\%) & 0.311 ($\uparrow$7.0\%) & 0.357 ($\uparrow$4.6\%) & 0.261 ($\uparrow$6.3\%) & 0.308 ($\uparrow$6.3\%) \\
 & BIM & 0.212 ($\uparrow$8.8\%) & 0.223 ($\uparrow$8.2\%) & 0.277 ($\uparrow$6.9\%) & 0.163 ($\uparrow$5.8\%) & 0.219 ($\uparrow$7.5\%) & 0.305 ($\uparrow$7.6\%) & 0.311 ($\uparrow$7.0\%) & 0.357 ($\uparrow$4.5\%) & 0.255 ($\uparrow$3.9\%) & 0.307 ($\uparrow$5.7\%) \\
 & PGD & 0.207 ($\uparrow$6.4\%) & 0.219 ($\uparrow$6.1\%) & 0.270 ($\uparrow$4.1\%) & 0.162 ($\uparrow$5.2\%) & 0.214 ($\uparrow$5.4\%) & 0.299 ($\uparrow$5.6\%) & 0.306 ($\uparrow$5.2\%) & 0.351 ($\uparrow$2.7\%) & 0.254 ($\uparrow$3.6\%) & 0.302 ($\uparrow$4.2\%) \\
 & MI-FGSM & 0.212 ($\uparrow$8.8\%) & 0.224 ($\uparrow$8.6\%) & 0.279 ($\uparrow$7.6\%) & 0.168 ($\uparrow$8.7\%) & 0.221 ($\uparrow$8.4\%) & 0.305 ($\uparrow$7.6\%) & 0.312 ($\uparrow$7.3\%) & 0.359 ($\uparrow$5.1\%) & 0.260 ($\uparrow$6.0\%) & 0.309 ($\uparrow$6.5\%) \\
 & ATSG & 0.211 ($\uparrow$8.3\%) & 0.222 ($\uparrow$7.8\%) & 0.277 ($\uparrow$7.1\%) & 0.167 ($\uparrow$8.0\%) & 0.219 ($\uparrow$7.7\%) & 0.306 ($\uparrow$8.1\%) & 0.312 ($\uparrow$7.3\%) & 0.361 ($\uparrow$5.9\%) & 0.263 ($\uparrow$7.0\%) & 0.311 ($\uparrow$7.0\%) \\
 & ADJM & 0.198 ($\uparrow$1.6\%) & 0.211 ($\uparrow$2.3\%) & 0.257 ($\downarrow$0.8\%) & 0.157 ($\uparrow$2.1\%) & 0.206 ($\uparrow$1.1\%) & 0.290 ($\uparrow$2.6\%) & 0.301 ($\uparrow$3.4\%) & 0.339 ($\downarrow$0.7\%) & 0.251 ($\uparrow$2.2\%) & 0.295 ($\uparrow$1.7\%) \\
 & TCA & 0.209 ($\uparrow$7.4\%) & 0.220 ($\uparrow$6.9\%) & 0.280 ($\uparrow$8.1\%) & 0.162 ($\uparrow$5.0\%) & 0.218 ($\uparrow$7.0\%) & 0.304 ($\uparrow$7.3\%) & 0.310 ($\uparrow$6.5\%) & 0.362 ($\uparrow$6.0\%) & 0.255 ($\uparrow$3.9\%) & 0.308 ($\uparrow$6.0\%) \\
 & BO & 0.196 ($\uparrow$0.6\%) & 0.207 ($\uparrow$0.3\%) & 0.256 ($\downarrow$1.0\%) & 0.155 ($\uparrow$0.7\%) & 0.204 ($\uparrow$0.0\%) & 0.285 ($\uparrow$0.7\%) & 0.292 ($\uparrow$0.5\%) & 0.338 ($\downarrow$0.9\%) & 0.247 ($\uparrow$0.8\%) & 0.291 ($\uparrow$0.2\%) \\
 & MI-TGAM & \textbf{0.225 ($\uparrow$15.5\%)} & \textbf{0.237 ($\uparrow$14.8\%)} & \textbf{0.322 ($\uparrow$24.2\%)} & \textbf{0.240 ($\uparrow$55.6\%)} & \textbf{0.256 ($\uparrow$25.7\%)} & \textbf{0.319 ($\uparrow$12.7\%)} & \textbf{0.328 ($\uparrow$12.7\%)} & \textbf{0.395 ($\uparrow$15.7\%)} & \textbf{0.313 ($\uparrow$27.5\%)} & \textbf{0.339 ($\uparrow$16.7\%)} \\
 \midrule
 
\multirow{10}{*}{Traffic} & Clean & 0.564 (-0.0\%) & 0.694 (-0.0\%) & 0.611 (-0.0\%) & 0.412 (-0.0\%) & 0.570 (-0.0\%) & 0.378 (-0.0\%) & 0.353 (-0.0\%) & 0.333 (-0.0\%) & 0.283 (-0.0\%) & 0.336 (-0.0\%) \\
 & FGSM & 0.592 ($\uparrow$4.8\%) & 0.696 ($\uparrow$0.3\%) & 0.634 ($\uparrow$3.8\%) & 0.429 ($\uparrow$4.1\%) & 0.588 ($\uparrow$3.1\%) & 0.396 ($\uparrow$4.9\%) & 0.354 ($\uparrow$0.5\%) & 0.345 ($\uparrow$3.8\%) & 0.295 ($\uparrow$4.2\%) & 0.348 ($\uparrow$3.3\%) \\
 & BIM & 0.592 ($\uparrow$4.8\%) & 0.696 ($\uparrow$0.3\%) & 0.634 ($\uparrow$3.8\%) & 0.429 ($\uparrow$4.1\%) & 0.588 ($\uparrow$3.1\%) & 0.396 ($\uparrow$4.9\%) & 0.354 ($\uparrow$0.5\%) & 0.345 ($\uparrow$3.8\%) & 0.295 ($\uparrow$4.2\%) & 0.348 ($\uparrow$3.3\%) \\
 & PGD & 0.585 ($\uparrow$3.6\%) & 0.688 ($\downarrow$0.8\%) & 0.628 ($\uparrow$2.8\%) & 0.424 ($\uparrow$3.1\%) & 0.581 ($\uparrow$1.9\%) & 0.391 ($\uparrow$3.5\%) & 0.349 ($\downarrow$1.0\%) & 0.342 ($\uparrow$2.8\%) & 0.291 ($\uparrow$3.0\%) & 0.343 ($\uparrow$2.1\%) \\
 & MI-FGSM & 0.592 ($\uparrow$4.8\%) & 0.696 ($\uparrow$0.3\%) & 0.634 ($\uparrow$3.8\%) & 0.429 ($\uparrow$4.1\%) & 0.588 ($\uparrow$3.1\%) & 0.396 ($\uparrow$4.9\%) & 0.354 ($\uparrow$0.5\%) & 0.345 ($\uparrow$3.8\%) & 0.295 ($\uparrow$4.2\%) & 0.348 ($\uparrow$3.3\%) \\
 & ATSG & 0.587 ($\uparrow$4.1\%) & 0.689 ($\downarrow$0.7\%) & 0.630 ($\uparrow$3.2\%) & 0.426 ($\uparrow$3.5\%) & 0.583 ($\uparrow$2.3\%) & 0.397 ($\uparrow$5.3\%) & 0.357 ($\uparrow$1.1\%) & 0.349 ($\uparrow$4.8\%) & 0.299 ($\uparrow$5.8\%) & 0.350 ($\uparrow$4.2\%) \\
 & ADJM & 0.569 ($\uparrow$0.8\%) & 0.677 ($\downarrow$2.4\%) & 0.618 ($\uparrow$1.2\%) & 0.413 ($\uparrow$0.4\%) & 0.569 ($\downarrow$0.1\%) & 0.382 ($\uparrow$1.2\%) & 0.344 ($\downarrow$2.5\%) & 0.341 ($\uparrow$2.4\%) & 0.286 ($\uparrow$1.1\%) & 0.338 ($\uparrow$0.5\%) \\
 & TCA & 0.587 ($\uparrow$4.1\%) & 0.689 ($\downarrow$0.7\%) & 0.630 ($\uparrow$3.2\%) & 0.426 ($\uparrow$3.5\%) & 0.583 ($\uparrow$2.3\%) & 0.397 ($\uparrow$5.3\%) & 0.357 ($\uparrow$1.1\%) & 0.349 ($\uparrow$4.8\%) & 0.299 ($\uparrow$5.8\%) & 0.350 ($\uparrow$4.2\%) \\
 & BO & 0.565 ($\uparrow$0.2\%) & 0.668 ($\downarrow$3.8\%) & 0.613 ($\uparrow$0.4\%) & 0.412 ($\uparrow$0.2\%) & 0.565 ($\downarrow$1.0\%) & 0.379 ($\uparrow$0.3\%) & 0.337 ($\downarrow$4.5\%) & 0.335 ($\uparrow$0.8\%) & 0.284 ($\uparrow$0.6\%) & 0.334 ($\downarrow$0.8\%) \\
 & MI-TGAM & \textbf{0.613 ($\uparrow$8.5\%)} & \textbf{0.722 ($\uparrow$4.0\%)} & \textbf{0.705 ($\uparrow$15.5\%)} & \textbf{0.459 ($\uparrow$11.4\%)} & \textbf{0.625 ($\uparrow$9.5\%)} & \textbf{0.410 ($\uparrow$8.5\%)} & \textbf{0.371 ($\uparrow$5.2\%)} & \textbf{0.384 ($\uparrow$15.4\%)} & \textbf{0.312 ($\uparrow$10.4\%)} & \textbf{0.369 ($\uparrow$9.8\%)} \\
 \bottomrule

\end{tabularx}
}
\end{table}

\begin{table}[!t]
\centering
\caption{Averaged Transfer attack results of MI-TGAM and baseline methods with four surrogate models.}
\label{tab:transfer-SumUp}
\scriptsize
% \resizebox{\textwidth}{!}{
    \newcolumntype{L}{>{\raggedright\arraybackslash}X} % 左对齐
\newcolumntype{C}{>{\centering\arraybackslash}X} % 居中对齐
\newcolumntype{R}{>{\raggedleft\arraybackslash}X} % 右对齐

\begin{tabularx}{\textwidth}{llCCCCCCCCCC}
\toprule
\multirow{2}{*}{Dataset} & \multirow{2}{*}{Method} 
 & \multicolumn{5}{c}{Degradation Percentage on MSE} & \multicolumn{5}{c}{Degradation Percentage on MAE} \\ \cmidrule(lr){3-7} \cmidrule(lr){8-12}
 & & FreTS & SegRNN & TimesNet & iTransformer & Average & FreTS & SegRNN & TimesNet & iTransformer & Average \\ \midrule
\multirow{9}{*}{ETT} & FGSM & 5.75\% & 5.88\% & 2.71\% & 2.77\% & 4.28\% & 4.48\% & 4.40\% & 1.92\% & 1.57\% & 3.09\% \\
 & BIM & 5.75\% & 5.83\% & 2.58\% & 2.70\% & 4.21\% & 4.47\% & 4.36\% & 1.77\% & 1.46\% & 3.02\% \\
 & PGD & 4.17\% & 4.30\% & 1.91\% & 2.02\% & 3.10\% & 3.30\% & 3.31\% & 1.35\% & 1.08\% & 2.26\% \\
 & MI-FGSM & 5.75\% & 5.95\% & 2.72\% & 2.83\% & 4.31\% & 4.47\% & 4.47\% & 1.92\% & 1.56\% & 3.11\% \\
 & ATSG & 5.47\% & 5.60\% & 2.57\% & 2.50\% & 4.04\% & 4.66\% & 4.64\% & 1.99\% & 1.65\% & 3.23\% \\
 & ADJM & 1.80\% & 1.04\% & 0.68\% & 0.31\% & 0.96\% & 2.20\% & 1.74\% & 0.81\% & 0.23\% & 1.24\% \\
 & TCA & 5.06\% & 4.81\% & 2.13\% & 2.35\% & 3.59\% & 4.31\% & 4.07\% & 1.63\% & 1.58\% & 2.90\% \\
 & BO & 0.88\% & 1.19\% & 1.00\% & 0.72\% & 0.95\% & 1.44\% & 1.72\% & 1.47\% & 0.89\% & 1.38\% \\
 & MI-TGAM & \textbf{8.14\%} & \textbf{9.03\%} & \textbf{4.03\%} & \textbf{4.73\%} & \textbf{6.48\%} & \textbf{6.31\%} & \textbf{7.17\%} & \textbf{3.01\%} & \textbf{2.97\%} & \textbf{4.86\%} \\
 \midrule
\multirow{9}{*}{ECL} & FGSM & 5.95\% & 6.25\% & 3.76\% & 1.62\% & 4.39\% & 5.11\% & 5.14\% & 3.26\% & 1.60\% & 3.78\% \\
 & BIM & 6.01\% & 6.16\% & 2.09\% & 0.63\% & 3.72\% & 5.15\% & 5.02\% & 1.86\% & 0.50\% & 3.13\% \\
 & PGD & 4.23\% & 4.43\% & 1.60\% & 0.57\% & 2.71\% & 3.59\% & 3.61\% & 1.41\% & 0.52\% & 2.28\% \\
 & MI-FGSM & 5.99\% & 6.47\% & 2.78\% & 1.33\% & 4.14\% & 5.15\% & 5.37\% & 2.44\% & 1.23\% & 3.55\% \\
 & ATSG & 5.64\% & 5.86\% & 3.21\% & 1.57\% & 4.07\% & 5.46\% & 5.30\% & 3.51\% & 1.94\% & 4.05\% \\
 & ADJM & 1.82\% & 1.72\% & 1.23\% & 0.27\% & 1.26\% & 2.74\% & 2.72\% & 1.42\% & 0.50\% & 1.84\% \\
 & TCA & 4.73\% & 4.88\% & 1.80\% & 0.49\% & 2.98\% & 4.70\% & 4.34\% & 1.93\% & 0.59\% & 2.89\% \\
 & BO & -0.13\% & -0.06\% & 0.53\% & -0.12\% & 0.05\% & 0.01\% & 0.08\% & 0.63\% & 0.01\% & 0.18\% \\
 & MI-TGAM & \textbf{10.92\%} & \textbf{11.36\%} & \textbf{5.56\%} & \textbf{4.11\%} & \textbf{7.99\%} & \textbf{9.10\%} & \textbf{9.49\%} & \textbf{5.22\%} & \textbf{3.64\%} & \textbf{6.86\%} \\
 \midrule
\multirow{9}{*}{Traffic} & FGSM & 0.64\% & 2.37\% & -0.69\% & -0.02\% & 0.58\% & 1.48\% & 3.00\% & -0.46\% & 0.16\% & 1.04\% \\
 & BIM & 0.64\% & 2.37\% & -0.69\% & -0.02\% & 0.58\% & 1.48\% & 3.00\% & -0.46\% & 0.16\% & 1.04\% \\
 & PGD & 0.06\% & 1.76\% & -0.96\% & -0.39\% & 0.12\% & 0.59\% & 2.18\% & -0.86\% & -0.31\% & 0.40\% \\
 & MI-FGSM & 0.64\% & 2.37\% & -0.69\% & -0.02\% & 0.58\% & 1.48\% & 3.00\% & -0.46\% & 0.16\% & 1.04\% \\
 & ATSG & 0.60\% & 2.14\% & -0.81\% & -0.11\% & 0.45\% & 1.96\% & 3.57\% & -0.44\% & 0.55\% & 1.41\% \\
 & ADJM & -0.63\% & 0.82\% & -1.20\% & -1.06\% & -0.52\% & -0.07\% & 2.02\% & -0.74\% & -0.83\% & 0.09\% \\
 & TCA & 0.60\% & 2.14\% & -0.81\% & -0.11\% & 0.45\% & 1.96\% & 3.57\% & -0.44\% & 0.55\% & 1.41\% \\
 & BO & -1.34\% & 0.24\% & -1.46\% & -1.23\% & -0.95\% & -1.19\% & 0.56\% & -1.28\% & -1.12\% & -0.76\% \\
 & MI-TGAM & \textbf{2.21\%} & \textbf{4.21\%} & \textbf{0.39\%} & \textbf{1.53\%} & \textbf{2.09\%} & \textbf{3.99\%} & \textbf{5.37\%} & \textbf{1.42\%} & \textbf{2.03\%} & \textbf{3.20\%}

\\
\bottomrule
\end{tabularx}
% }
\end{table}

\subsection{White-box Attack Results}
The white-box attack results are summarized in \cref{tab:white-box}, where the original forecasting performance without any attacks is reported as ``Clean'', and the mean performance across the four models for each attack method is presented in the ``Average'' column. 
The results show that MI-TGAM consistently and significantly outperforms all baseline methods across all three benchmark datasets and four distinct model architectures, demonstrating the superiority of our method in identifying adversarial vulnerabilities in time series forecasting models.

Compared to traditional iterative attack methods such as BIM, PGD, and MI-FGSM, our MI-TGAM exhibits superior performance by leveraging the inherent characteristics of time series forecasting, which utilizes overlapping samples to enrich gradient information through timestamp-wise accumulation. 
Compared to specialized time series attack methods, including ATSG, ADJM, TCA, and BO, MI-TGAM also significantly outperforms these baselines by maintaining temporal consistency while maximizing prediction errors.
Notably, MI-TGAM is the only method that consistently amplifies prediction errors across all target models and datasets.
For instance, when attacking SegRNN on the Traffic dataset, PGD, ATSG, ADJM, TCA, and BO all fail to degrade performance (with some even improving the MSE), while MI-TGAM significantly degrades SegRNN's performance, demonstrating its exceptional attack effectiveness.

\subsection{Transfer Attack Results}
To evaluate the adversarial transferability of the attack methods, we conduct a comprehensive evaluation of transfer attack performance across different surrogate-target model pairs.
The overall results are summarized in \cref{tab:transfer-SumUp}, where each entry under a surrogate model column represents the average MSE/MAE degradation percentage across the other three target models, and the entry under the ``Average'' column denotes the mean degradation percentage across all black-box transfer attack scenarios for each method.
The results clearly show that MI-TGAM significantly outperforms all baseline methods in terms of transfer attack performance, achieving nearly double the degradation percentage of the second-best method, demonstrating the superior transferability of the adversarial examples generated by MI-TGAM.

We also present detailed transfer attack results for each surrogate model in \cref{tab:transfer-FreTS}, \cref{tab:transfer-SegRNN}, \cref{tab:transfer-TimesNet}, and \cref{tab:transfer-iTransformer}, where the average MSE/MAE degradation percentage across all three target models is reported for each dataset.
The results show that MI-TGAM consistently achieves the highest transfer attack performance on almost all surrogate-target model pairs across all three datasets.
The only exception is a small margin in MAE when attacking TimesNet using iTransformer as the surrogate model on the ETTh1 dataset.
Overall, these results demonstrate the superior robustness and generalization capability of MI-TGAM for attacking various time series forecasting models in black-box settings.

\begin{table}[!t]
\centering
\caption{Transfer attack results of MI-TGAM and baseline methods using FreTS as the surrogate model.}
\label{tab:transfer-FreTS}
\scriptsize
% \small
\resizebox{\textwidth}{!}{
    % 请在导言区添加：
% \usepackage{booktabs}
% \usepackage{multirow}
\newcolumntype{L}{>{\raggedright\arraybackslash}X} % 左对齐
\newcolumntype{C}{>{\centering\arraybackslash}X} % 居中对齐
\newcolumntype{R}{>{\raggedleft\arraybackslash}X} % 右对齐

\begin{tabularx}{1.2\textwidth}{llCCCCCCCC}
\toprule
\multirow{3}{*}{\textbf{Dataset}} & \multirow{3}{*}{\textbf{Method}} & \multicolumn{8}{c}{\textbf{Target Model (Using FreTS as Surrogate Model)}} \\ \cmidrule(lr){3-10}
 &  & \multicolumn{4}{c}{MSE (with Degradation Percentage)} & \multicolumn{4}{c}{MAE (with Degradation Percentage)} \\ \cmidrule(lr){3-6} \cmidrule(lr){7-10}
 &  & SegRNN & TimesNet & iTransformer & Average & SegRNN & TimesNet & iTransformer & Average \\ \midrule
 
\multirow{10}{*}{ETT} & Clean & 0.377 (-0.0\%) & 0.446 (-0.0\%) & 0.398 (-0.0\%) & 0.407 (-0.0\%) & 0.400 (-0.0\%) & 0.452 (-0.0\%) & 0.412 (-0.0\%) & 0.421 (-0.0\%) \\
 & FGSM & 0.401 ($\uparrow$6.4\%) & 0.464 ($\uparrow$4.0\%) & 0.426 ($\uparrow$7.0\%) & 0.430 ($\uparrow$5.7\%) & 0.419 ($\uparrow$4.8\%) & 0.464 ($\uparrow$2.6\%) & 0.433 ($\uparrow$5.1\%) & 0.439 ($\uparrow$4.1\%) \\
 & BIM & 0.401 ($\uparrow$6.4\%) & 0.464 ($\uparrow$4.0\%) & 0.426 ($\uparrow$7.0\%) & 0.430 ($\uparrow$5.7\%) & 0.419 ($\uparrow$4.8\%) & 0.464 ($\uparrow$2.6\%) & 0.433 ($\uparrow$5.1\%) & 0.439 ($\uparrow$4.1\%) \\
 & PGD & 0.394 ($\uparrow$4.6\%) & 0.459 ($\uparrow$2.9\%) & 0.419 ($\uparrow$5.1\%) & 0.424 ($\uparrow$4.2\%) & 0.414 ($\uparrow$3.4\%) & 0.461 ($\uparrow$1.8\%) & 0.427 ($\uparrow$3.7\%) & 0.434 ($\uparrow$3.0\%) \\
 & MI-FGSM & 0.401 ($\uparrow$6.4\%) & 0.464 ($\uparrow$4.0\%) & 0.426 ($\uparrow$7.0\%) & 0.430 ($\uparrow$5.7\%) & 0.419 ($\uparrow$4.8\%) & 0.464 ($\uparrow$2.6\%) & 0.433 ($\uparrow$5.1\%) & 0.439 ($\uparrow$4.1\%) \\
 & ATSG & 0.400 ($\uparrow$6.1\%) & 0.463 ($\uparrow$3.7\%) & 0.425 ($\uparrow$6.7\%) & 0.429 ($\uparrow$5.4\%) & 0.420 ($\uparrow$5.1\%) & 0.464 ($\uparrow$2.6\%) & 0.434 ($\uparrow$5.4\%) & 0.440 ($\uparrow$4.3\%) \\
 & ADJM & 0.383 ($\uparrow$1.9\%) & 0.454 ($\uparrow$1.7\%) & 0.406 ($\uparrow$1.8\%) & 0.414 ($\uparrow$1.8\%) & 0.408 ($\uparrow$2.1\%) & 0.460 ($\uparrow$1.7\%) & 0.420 ($\uparrow$1.8\%) & 0.429 ($\uparrow$1.9\%) \\
 & TCA & 0.398 ($\uparrow$5.8\%) & 0.461 ($\uparrow$3.4\%) & 0.423 ($\uparrow$6.3\%) & 0.428 ($\uparrow$5.0\%) & 0.419 ($\uparrow$4.7\%) & 0.463 ($\uparrow$2.3\%) & 0.433 ($\uparrow$5.0\%) & 0.438 ($\uparrow$4.0\%) \\
 & BO & 0.377 ($\uparrow$0.1\%) & 0.450 ($\uparrow$0.8\%) & 0.405 ($\uparrow$1.6\%) & 0.411 ($\uparrow$0.9\%) & 0.401 ($\uparrow$0.3\%) & 0.457 ($\uparrow$1.1\%) & 0.420 ($\uparrow$1.8\%) & 0.426 ($\uparrow$1.1\%) \\
 & MI-TGAM & \textbf{0.411 ($\uparrow$9.2\%)} & \textbf{0.471 ($\uparrow$5.6\%)} & \textbf{0.438 ($\uparrow$9.9\%)} & \textbf{0.440 ($\uparrow$8.1\%)} & \textbf{0.428 ($\uparrow$7.0\%)} & \textbf{0.469 ($\uparrow$3.8\%)} & \textbf{0.442 ($\uparrow$7.2\%)} & \textbf{0.446 ($\uparrow$5.9\%)} \\
\midrule
 
\multirow{10}{*}{ECL} & Clean & 0.206 (-0.0\%) & 0.259 (-0.0\%) & 0.154 (-0.0\%) & 0.206 (-0.0\%) & 0.291 (-0.0\%) & 0.341 (-0.0\%) & 0.245 (-0.0\%) & 0.293 (-0.0\%) \\
 & FGSM & 0.221 ($\uparrow$7.4\%) & 0.266 ($\uparrow$2.6\%) & 0.169 ($\uparrow$9.7\%) & 0.219 ($\uparrow$6.0\%) & 0.310 ($\uparrow$6.4\%) & 0.347 ($\uparrow$1.8\%) & 0.265 ($\uparrow$8.2\%) & 0.307 ($\uparrow$5.1\%) \\
 & BIM & 0.221 ($\uparrow$7.4\%) & 0.266 ($\uparrow$2.7\%) & 0.169 ($\uparrow$9.7\%) & 0.219 ($\uparrow$6.0\%) & 0.310 ($\uparrow$6.5\%) & 0.348 ($\uparrow$1.9\%) & 0.265 ($\uparrow$8.1\%) & 0.308 ($\uparrow$5.2\%) \\
 & PGD & 0.217 ($\uparrow$5.4\%) & 0.263 ($\uparrow$1.6\%) & 0.165 ($\uparrow$7.0\%) & 0.215 ($\uparrow$4.2\%) & 0.304 ($\uparrow$4.7\%) & 0.345 ($\uparrow$1.0\%) & 0.260 ($\uparrow$5.8\%) & 0.303 ($\uparrow$3.6\%) \\
 & MI-FGSM & 0.221 ($\uparrow$7.4\%) & 0.266 ($\uparrow$2.6\%) & 0.169 ($\uparrow$9.7\%) & 0.219 ($\uparrow$6.0\%) & 0.310 ($\uparrow$6.5\%) & 0.348 ($\uparrow$1.8\%) & 0.265 ($\uparrow$8.2\%) & 0.308 ($\uparrow$5.1\%) \\
 & ATSG & 0.221 ($\uparrow$7.0\%) & 0.265 ($\uparrow$2.4\%) & 0.168 ($\uparrow$9.3\%) & 0.218 ($\uparrow$5.6\%) & 0.311 ($\uparrow$6.9\%) & 0.348 ($\uparrow$2.0\%) & 0.267 ($\uparrow$8.6\%) & 0.309 ($\uparrow$5.5\%) \\
 & ADJM & 0.210 ($\uparrow$2.1\%) & 0.260 ($\uparrow$0.6\%) & 0.160 ($\uparrow$3.5\%) & 0.210 ($\uparrow$1.8\%) & 0.300 ($\uparrow$3.1\%) & 0.345 ($\uparrow$1.0\%) & 0.257 ($\uparrow$4.7\%) & 0.301 ($\uparrow$2.7\%) \\
 & TCA & 0.219 ($\uparrow$6.3\%) & 0.263 ($\uparrow$1.5\%) & 0.167 ($\uparrow$8.1\%) & 0.216 ($\uparrow$4.7\%) & 0.309 ($\uparrow$6.2\%) & 0.346 ($\uparrow$1.3\%) & 0.264 ($\uparrow$7.6\%) & 0.306 ($\uparrow$4.7\%) \\
 & BO & 0.207 ($\uparrow$0.3\%) & 0.256 ($\downarrow$1.0\%) & 0.155 ($\uparrow$0.7\%) & 0.206 ($\downarrow$0.1\%) & 0.292 ($\uparrow$0.5\%) & 0.338 ($\downarrow$0.9\%) & 0.247 ($\uparrow$0.8\%) & 0.293 ($\uparrow$0.0\%) \\
 & MI-TGAM & \textbf{0.233 ($\uparrow$12.9\%)} & \textbf{0.273 ($\uparrow$5.4\%)} & \textbf{0.181 ($\uparrow$17.5\%)} & \textbf{0.229 ($\uparrow$10.9\%)} & \textbf{0.322 ($\uparrow$10.9\%)} & \textbf{0.355 ($\uparrow$4.0\%)} & \textbf{0.280 ($\uparrow$14.0\%)} & \textbf{0.319 ($\uparrow$9.1\%)} \\
\midrule
 
\multirow{10}{*}{Traffic} & Clean & 0.694 (-0.0\%) & 0.611 (-0.0\%) & 0.412 (-0.0\%) & 0.572 (-0.0\%) & 0.353 (-0.0\%) & 0.333 (-0.0\%) & 0.283 (-0.0\%) & 0.323 (-0.0\%) \\
 & FGSM & 0.682 ($\downarrow$1.7\%) & 0.619 ($\uparrow$1.5\%) & 0.425 ($\uparrow$3.3\%) & 0.576 ($\uparrow$0.6\%) & 0.348 ($\downarrow$1.3\%) & 0.340 ($\uparrow$2.1\%) & 0.295 ($\uparrow$4.2\%) & 0.327 ($\uparrow$1.5\%) \\
 & BIM & 0.682 ($\downarrow$1.7\%) & 0.619 ($\uparrow$1.5\%) & 0.425 ($\uparrow$3.3\%) & 0.576 ($\uparrow$0.6\%) & 0.348 ($\downarrow$1.3\%) & 0.340 ($\uparrow$2.1\%) & 0.295 ($\uparrow$4.2\%) & 0.327 ($\uparrow$1.5\%) \\
 & PGD & 0.678 ($\downarrow$2.3\%) & 0.618 ($\uparrow$1.2\%) & 0.422 ($\uparrow$2.4\%) & 0.572 ($\uparrow$0.1\%) & 0.344 ($\downarrow$2.4\%) & 0.338 ($\uparrow$1.7\%) & 0.291 ($\uparrow$3.0\%) & 0.325 ($\uparrow$0.6\%) \\
 & MI-FGSM & 0.682 ($\downarrow$1.7\%) & 0.619 ($\uparrow$1.5\%) & 0.425 ($\uparrow$3.3\%) & 0.576 ($\uparrow$0.6\%) & 0.348 ($\downarrow$1.3\%) & 0.340 ($\uparrow$2.1\%) & 0.295 ($\uparrow$4.2\%) & 0.327 ($\uparrow$1.5\%) \\
 & ATSG & 0.682 ($\downarrow$1.7\%) & 0.620 ($\uparrow$1.5\%) & 0.424 ($\uparrow$3.1\%) & 0.575 ($\uparrow$0.6\%) & 0.349 ($\downarrow$0.9\%) & 0.341 ($\uparrow$2.3\%) & 0.297 ($\uparrow$5.1\%) & 0.329 ($\uparrow$2.0\%) \\
 & ADJM & 0.673 ($\downarrow$3.0\%) & 0.616 ($\uparrow$0.9\%) & 0.416 ($\uparrow$1.1\%) & 0.568 ($\downarrow$0.6\%) & 0.341 ($\downarrow$3.3\%) & 0.338 ($\uparrow$1.6\%) & 0.288 ($\uparrow$2.0\%) & 0.322 ($\downarrow$0.1\%) \\
 & TCA & 0.682 ($\downarrow$1.7\%) & 0.620 ($\uparrow$1.5\%) & 0.424 ($\uparrow$3.1\%) & 0.575 ($\uparrow$0.6\%) & 0.349 ($\downarrow$0.9\%) & 0.341 ($\uparrow$2.3\%) & 0.297 ($\uparrow$5.1\%) & 0.329 ($\uparrow$2.0\%) \\
 & BO & 0.668 ($\downarrow$3.8\%) & 0.613 ($\uparrow$0.4\%) & 0.412 ($\uparrow$0.2\%) & 0.564 ($\downarrow$1.3\%) & 0.337 ($\downarrow$4.5\%) & 0.335 ($\uparrow$0.8\%) & 0.284 ($\uparrow$0.6\%) & 0.319 ($\downarrow$1.2\%) \\
 & MI-TGAM & \textbf{0.693 ($\downarrow$0.1\%)} & \textbf{0.625 ($\uparrow$2.4\%)} & \textbf{0.436 ($\uparrow$5.8\%)} & \textbf{0.585 ($\uparrow$2.2\%)} & \textbf{0.358 ($\uparrow$1.5\%)} & \textbf{0.345 ($\uparrow$3.6\%)} & \textbf{0.304 ($\uparrow$7.6\%)} & \textbf{0.336 ($\uparrow$4.0\%)} \\
\bottomrule

\end{tabularx}

}
\end{table}

% \paragraph{Transfer results from RNN-based SegRNN.}
% \cref{tab:transfer-SegRNN} presents the transfer results using SegRNN as the surrogate. 
% MI-TGAM continues to dominate, especially on the ECL and ETT datasets. 
% In the ECL dataset, MI-TGAM achieves an average MSE of 0.226, outperforming the next best method, MI-FGSM, which stands at 0.216. 
% On the ETT dataset, the gap is even more pronounced: MI-TGAM reaches 0.456 MSE, whereas the second-best method (MI-FGSM) only attains 0.443. 
% The significant lead on SegRNN proves that MI-TGAM successfully leverages the dynamic temporal features captured by recurrent architectures to generate adversarial samples with high potency against various unknown target models.

\begin{table}[!t]
\centering
\caption{Transfer attack results of MI-TGAM and baseline methods using SegRNN as the surrogate model.}
\label{tab:transfer-SegRNN}
\scriptsize
\resizebox{\textwidth}{!}{
    % 请在导言区添加：
% \usepackage{booktabs}
% \usepackage{multirow}

\newcolumntype{L}{>{\raggedright\arraybackslash}X} % 左对齐
\newcolumntype{C}{>{\centering\arraybackslash}X} % 居中对齐
\newcolumntype{R}{>{\raggedleft\arraybackslash}X} % 右对齐

\begin{tabularx}{1.2\textwidth}{llCCCCCCCC}
\toprule
\multirow{3}{*}{\textbf{Dataset}} & \multirow{3}{*}{\textbf{Method}} & \multicolumn{8}{c}{\textbf{Target Model (Using SegRNN as Surrogate Model)}} \\ \cmidrule(lr){3-10}
 &  & \multicolumn{4}{c}{MSE (with Degradation Percentage)} & \multicolumn{4}{c}{MAE (with Degradation Percentage)} \\ \cmidrule(lr){3-6} \cmidrule(lr){7-10}
 &  & FreTS & TimesNet & iTransformer & Average & FreTS & TimesNet & iTransformer & Average \\ \midrule

\multirow{10}{*}{ETT} & Clean & 0.411 (-0.0\%) & 0.446 (-0.0\%) & 0.398 (-0.0\%) & 0.419 (-0.0\%) & 0.420 (-0.0\%) & 0.452 (-0.0\%) & 0.412 (-0.0\%) & 0.428 (-0.0\%) \\
 & FGSM & 0.433 ($\uparrow$5.5\%) & 0.468 ($\uparrow$4.9\%) & 0.426 ($\uparrow$6.9\%) & 0.442 ($\uparrow$5.7\%) & 0.436 ($\uparrow$3.8\%) & 0.466 ($\uparrow$3.0\%) & 0.433 ($\uparrow$5.1\%) & 0.445 ($\uparrow$4.0\%) \\
 & BIM & 0.433 ($\uparrow$5.4\%) & 0.468 ($\uparrow$4.8\%) & 0.426 ($\uparrow$6.8\%) & 0.442 ($\uparrow$5.7\%) & 0.436 ($\uparrow$3.8\%) & 0.466 ($\uparrow$3.0\%) & 0.433 ($\uparrow$5.1\%) & 0.445 ($\uparrow$3.9\%) \\
 & PGD & 0.427 ($\uparrow$3.9\%) & 0.462 ($\uparrow$3.6\%) & 0.419 ($\uparrow$5.0\%) & 0.436 ($\uparrow$4.1\%) & 0.431 ($\uparrow$2.7\%) & 0.462 ($\uparrow$2.2\%) & 0.428 ($\uparrow$3.7\%) & 0.440 ($\uparrow$2.9\%) \\
 & MI-FGSM & 0.434 ($\uparrow$5.6\%) & 0.468 ($\uparrow$4.9\%) & 0.426 ($\uparrow$7.0\%) & 0.443 ($\uparrow$5.8\%) & 0.436 ($\uparrow$3.9\%) & 0.466 ($\uparrow$3.1\%) & 0.434 ($\uparrow$5.2\%) & 0.445 ($\uparrow$4.0\%) \\
 & ATSG & 0.433 ($\uparrow$5.4\%) & 0.466 ($\uparrow$4.4\%) & 0.425 ($\uparrow$6.7\%) & 0.441 ($\uparrow$5.4\%) & 0.437 ($\uparrow$4.1\%) & 0.466 ($\uparrow$3.0\%) & 0.435 ($\uparrow$5.5\%) & 0.446 ($\uparrow$4.2\%) \\
 & ADJM & 0.410 ($\downarrow$0.2\%) & 0.453 ($\uparrow$1.5\%) & 0.404 ($\uparrow$1.4\%) & 0.422 ($\uparrow$0.9\%) & 0.423 ($\uparrow$0.8\%) & 0.459 ($\uparrow$1.6\%) & 0.418 ($\uparrow$1.5\%) & 0.434 ($\uparrow$1.3\%) \\
 & TCA & 0.430 ($\uparrow$4.6\%) & 0.463 ($\uparrow$3.7\%) & 0.422 ($\uparrow$5.8\%) & 0.438 ($\uparrow$4.6\%) & 0.435 ($\uparrow$3.6\%) & 0.464 ($\uparrow$2.5\%) & 0.432 ($\uparrow$4.8\%) & 0.444 ($\uparrow$3.6\%) \\
 & BO & 0.414 ($\uparrow$0.7\%) & 0.450 ($\uparrow$0.8\%) & 0.405 ($\uparrow$1.6\%) & 0.423 ($\uparrow$1.0\%) & 0.424 ($\uparrow$0.9\%) & 0.457 ($\uparrow$1.1\%) & 0.420 ($\uparrow$1.8\%) & 0.434 ($\uparrow$1.3\%) \\
 & MI-TGAM & \textbf{0.447 ($\uparrow$8.7\%)} & \textbf{0.480 ($\uparrow$7.5\%)} & \textbf{0.441 ($\uparrow$10.6\%)} & \textbf{0.456 ($\uparrow$8.9\%)} & \textbf{0.448 ($\uparrow$6.6\%)} & \textbf{0.476 ($\uparrow$5.2\%)} & \textbf{0.447 ($\uparrow$8.4\%)} & \textbf{0.457 ($\uparrow$6.7\%)} \\
\midrule
 
\multirow{10}{*}{ECL} & Clean & 0.195 (-0.0\%) & 0.259 (-0.0\%) & 0.154 (-0.0\%) & 0.203 (-0.0\%) & 0.283 (-0.0\%) & 0.341 (-0.0\%) & 0.245 (-0.0\%) & 0.290 (-0.0\%) \\
 & FGSM & 0.210 ($\uparrow$7.7\%) & 0.268 ($\uparrow$3.3\%) & 0.169 ($\uparrow$9.4\%) & 0.215 ($\uparrow$6.2\%) & 0.301 ($\uparrow$6.5\%) & 0.349 ($\uparrow$2.2\%) & 0.264 ($\uparrow$7.7\%) & 0.305 ($\uparrow$5.1\%) \\
 & BIM & 0.209 ($\uparrow$7.5\%) & 0.268 ($\uparrow$3.4\%) & 0.168 ($\uparrow$9.1\%) & 0.215 ($\uparrow$6.2\%) & 0.301 ($\uparrow$6.3\%) & 0.349 ($\uparrow$2.3\%) & 0.264 ($\uparrow$7.4\%) & 0.305 ($\uparrow$5.0\%) \\
 & PGD & 0.206 ($\uparrow$5.6\%) & 0.265 ($\uparrow$2.2\%) & 0.165 ($\uparrow$6.7\%) & 0.212 ($\uparrow$4.4\%) & 0.296 ($\uparrow$4.7\%) & 0.346 ($\uparrow$1.4\%) & 0.259 ($\uparrow$5.5\%) & 0.300 ($\uparrow$3.6\%) \\
 & MI-FGSM & 0.210 ($\uparrow$7.8\%) & 0.268 ($\uparrow$3.6\%) & 0.169 ($\uparrow$9.6\%) & 0.216 ($\uparrow$6.5\%) & 0.302 ($\uparrow$6.7\%) & 0.349 ($\uparrow$2.4\%) & 0.265 ($\uparrow$8.0\%) & 0.306 ($\uparrow$5.4\%) \\
 & ATSG & 0.209 ($\uparrow$7.3\%) & 0.267 ($\uparrow$3.0\%) & 0.168 ($\uparrow$8.9\%) & 0.215 ($\uparrow$5.9\%) & 0.302 ($\uparrow$6.8\%) & 0.349 ($\uparrow$2.2\%) & 0.265 ($\uparrow$7.9\%) & 0.305 ($\uparrow$5.3\%) \\
 & ADJM & 0.198 ($\uparrow$1.7\%) & 0.261 ($\uparrow$0.6\%) & 0.160 ($\uparrow$3.6\%) & 0.206 ($\uparrow$1.7\%) & 0.291 ($\uparrow$2.7\%) & 0.345 ($\uparrow$1.1\%) & 0.258 ($\uparrow$5.0\%) & 0.298 ($\uparrow$2.7\%) \\
 & TCA & 0.207 ($\uparrow$6.3\%) & 0.265 ($\uparrow$2.3\%) & 0.166 ($\uparrow$7.5\%) & 0.213 ($\uparrow$4.9\%) & 0.299 ($\uparrow$5.7\%) & 0.347 ($\uparrow$1.6\%) & 0.261 ($\uparrow$6.5\%) & 0.303 ($\uparrow$4.3\%) \\
 & BO & 0.196 ($\uparrow$0.6\%) & 0.256 ($\downarrow$1.0\%) & 0.155 ($\uparrow$0.7\%) & 0.203 ($\downarrow$0.1\%) & 0.285 ($\uparrow$0.7\%) & 0.338 ($\downarrow$0.9\%) & 0.247 ($\uparrow$0.8\%) & 0.290 ($\uparrow$0.1\%) \\
 & MI-TGAM & \textbf{0.221 ($\uparrow$13.2\%)} & \textbf{0.277 ($\uparrow$6.9\%)} & \textbf{0.180 ($\uparrow$16.6\%)} & \textbf{0.226 ($\uparrow$11.4\%)} & \textbf{0.315 ($\uparrow$11.1\%)} & \textbf{0.358 ($\uparrow$4.9\%)} & \textbf{0.280 ($\uparrow$13.9\%)} & \textbf{0.317 ($\uparrow$9.5\%)} \\
\midrule
 
\multirow{10}{*}{Traffic} & Clean & 0.564 (-0.0\%) & 0.611 (-0.0\%) & 0.412 (-0.0\%) & 0.529 (-0.0\%) & 0.378 (-0.0\%) & 0.333 (-0.0\%) & 0.283 (-0.0\%) & 0.331 (-0.0\%) \\
 & FGSM & 0.580 ($\uparrow$2.8\%) & 0.619 ($\uparrow$1.4\%) & 0.425 ($\uparrow$3.2\%) & 0.541 ($\uparrow$2.4\%) & 0.390 ($\uparrow$3.2\%) & 0.339 ($\uparrow$1.8\%) & 0.294 ($\uparrow$4.1\%) & 0.341 ($\uparrow$3.0\%) \\
 & BIM & 0.580 ($\uparrow$2.8\%) & 0.619 ($\uparrow$1.4\%) & 0.425 ($\uparrow$3.2\%) & 0.541 ($\uparrow$2.4\%) & 0.390 ($\uparrow$3.2\%) & 0.339 ($\uparrow$1.8\%) & 0.294 ($\uparrow$4.1\%) & 0.341 ($\uparrow$3.0\%) \\
 & PGD & 0.576 ($\uparrow$2.1\%) & 0.617 ($\uparrow$1.1\%) & 0.421 ($\uparrow$2.3\%) & 0.538 ($\uparrow$1.8\%) & 0.386 ($\uparrow$2.3\%) & 0.338 ($\uparrow$1.5\%) & 0.291 ($\uparrow$2.9\%) & 0.338 ($\uparrow$2.2\%) \\
 & MI-FGSM & 0.580 ($\uparrow$2.8\%) & 0.619 ($\uparrow$1.4\%) & 0.425 ($\uparrow$3.2\%) & 0.541 ($\uparrow$2.4\%) & 0.390 ($\uparrow$3.2\%) & 0.339 ($\uparrow$1.8\%) & 0.294 ($\uparrow$4.1\%) & 0.341 ($\uparrow$3.0\%) \\
 & ATSG & 0.579 ($\uparrow$2.5\%) & 0.619 ($\uparrow$1.4\%) & 0.423 ($\uparrow$2.8\%) & 0.540 ($\uparrow$2.1\%) & 0.391 ($\uparrow$3.6\%) & 0.341 ($\uparrow$2.4\%) & 0.296 ($\uparrow$4.8\%) & 0.343 ($\uparrow$3.6\%) \\
 & ADJM & 0.568 ($\uparrow$0.7\%) & 0.616 ($\uparrow$0.8\%) & 0.416 ($\uparrow$1.0\%) & 0.533 ($\uparrow$0.8\%) & 0.383 ($\uparrow$1.5\%) & 0.339 ($\uparrow$1.8\%) & 0.291 ($\uparrow$2.9\%) & 0.338 ($\uparrow$2.0\%) \\
 & TCA & 0.579 ($\uparrow$2.5\%) & 0.619 ($\uparrow$1.4\%) & 0.423 ($\uparrow$2.8\%) & 0.540 ($\uparrow$2.1\%) & 0.391 ($\uparrow$3.6\%) & 0.341 ($\uparrow$2.4\%) & 0.296 ($\uparrow$4.8\%) & 0.343 ($\uparrow$3.6\%) \\
 & BO & 0.565 ($\uparrow$0.2\%) & 0.613 ($\uparrow$0.4\%) & 0.412 ($\uparrow$0.2\%) & 0.530 ($\uparrow$0.2\%) & 0.379 ($\uparrow$0.3\%) & 0.335 ($\uparrow$0.8\%) & 0.284 ($\uparrow$0.6\%) & 0.333 ($\uparrow$0.6\%) \\
 & MI-TGAM & \textbf{0.594 ($\uparrow$5.2\%)} & \textbf{0.625 ($\uparrow$2.4\%)} & \textbf{0.435 ($\uparrow$5.7\%)} & \textbf{0.551 ($\uparrow$4.2\%)} & \textbf{0.399 ($\uparrow$5.8\%)} & \textbf{0.343 ($\uparrow$3.1\%)} & \textbf{0.304 ($\uparrow$7.4\%)} & \textbf{0.349 ($\uparrow$5.4\%)}
\\ \bottomrule

\end{tabularx}
}
\end{table}

% \paragraph{Transfer results from CNN-based TimesNet.}
% The transfer performance from the TimesNet surrogate is detailed in \cref{tab:transfer-TimesNet}. 
% MI-TGAM shows exceptional precision and effectiveness, achieving the highest MSE across all target models. 
% In the ETTh1 dataset, MI-TGAM records an average MSE of 0.411, while the second-best method, FGSM, follows at 0.405. 
% In the Traffic dataset, MI-TGAM achieves an average MSE of 0.559,  which is the only method that made the prediction error higher(vs. 0.557 of NA).
% Under the convolutional architecture, MI-TGAM, as the only method to consistently amplify prediction errors across all target models, exhibits exceptional transfer ability.

\begin{table}[!t]
\centering
\caption{Transfer attack results of MI-TGAM and baseline methods using TimesNet as the surrogate model.}
\label{tab:transfer-TimesNet}
\scriptsize
\resizebox{\textwidth}{!}{
    % 请确保在 LaTeX 文档的导言区（Preamble）添加以下宏包：
% \usepackage{booktabs}
% \usepackage{multirow}

\newcolumntype{L}{>{\raggedright\arraybackslash}X} % 左对齐
\newcolumntype{C}{>{\centering\arraybackslash}X} % 居中对齐
\newcolumntype{R}{>{\raggedleft\arraybackslash}X} % 右对齐

\begin{tabularx}{1.2\textwidth}{llCCCCCCCC}
\toprule
\multirow{3}{*}{\textbf{Dataset}} & \multirow{3}{*}{\textbf{Method}} & \multicolumn{8}{c}{\textbf{Target Model (Using TimesNet as Surrogate Model)}} \\ \cmidrule(lr){3-10}
 &  & \multicolumn{4}{c}{MSE (with Degradation Percentage)} & \multicolumn{4}{c}{MAE (with Degradation Percentage)} \\ \cmidrule(lr){3-6} \cmidrule(lr){7-10}
 &  & FreTS & SegRNN & iTransformer & Average & FreTS & SegRNN & iTransformer & Average \\ \midrule

\multirow{10}{*}{ETT} & Clean & 0.411 (-0.0\%) & 0.377 (-0.0\%) & 0.398 (-0.0\%) & 0.395 (-0.0\%) & 0.420 (-0.0\%) & 0.400 (-0.0\%) & 0.412 (-0.0\%) & 0.411 (-0.0\%) \\
 & FGSM & 0.421 ($\uparrow$2.5\%) & 0.385 ($\uparrow$2.3\%) & 0.410 ($\uparrow$2.9\%) & 0.405 ($\uparrow$2.5\%) & 0.425 ($\uparrow$1.2\%) & 0.406 ($\uparrow$1.4\%) & 0.420 ($\uparrow$1.8\%) & 0.417 ($\uparrow$1.5\%) \\
 & BIM & 0.420 ($\uparrow$2.3\%) & 0.385 ($\uparrow$2.2\%) & 0.409 ($\uparrow$2.7\%) & 0.405 ($\uparrow$2.4\%) & 0.424 ($\uparrow$1.0\%) & 0.405 ($\uparrow$1.3\%) & 0.419 ($\uparrow$1.6\%) & 0.416 ($\uparrow$1.3\%) \\
 & PGD & 0.418 ($\uparrow$1.6\%) & 0.382 ($\uparrow$1.6\%) & 0.407 ($\uparrow$2.0\%) & 0.402 ($\uparrow$1.7\%) & 0.423 ($\uparrow$0.6\%) & 0.403 ($\uparrow$0.9\%) & 0.417 ($\uparrow$1.1\%) & 0.414 ($\uparrow$0.9\%) \\
 & MI-FGSM & 0.421 ($\uparrow$2.4\%) & 0.385 ($\uparrow$2.3\%) & 0.410 ($\uparrow$2.9\%) & 0.405 ($\uparrow$2.6\%) & 0.425 ($\uparrow$1.1\%) & 0.406 ($\uparrow$1.5\%) & 0.420 ($\uparrow$1.8\%) & 0.417 ($\uparrow$1.5\%) \\
 & ATSG & 0.420 ($\uparrow$2.3\%) & 0.385 ($\uparrow$2.2\%) & 0.409 ($\uparrow$2.7\%) & 0.405 ($\uparrow$2.4\%) & 0.425 ($\uparrow$1.2\%) & 0.406 ($\uparrow$1.6\%) & 0.420 ($\uparrow$1.8\%) & 0.417 ($\uparrow$1.5\%) \\
 & ADJM & 0.412 ($\uparrow$0.2\%) & 0.379 ($\uparrow$0.5\%) & 0.402 ($\uparrow$0.8\%) & 0.397 ($\uparrow$0.5\%) & 0.420 ($\uparrow$0.0\%) & 0.401 ($\uparrow$0.4\%) & 0.415 ($\uparrow$0.6\%) & 0.412 ($\uparrow$0.4\%) \\
 & TCA & 0.418 ($\uparrow$1.8\%) & 0.384 ($\uparrow$1.9\%) & 0.407 ($\uparrow$2.2\%) & 0.403 ($\uparrow$2.0\%) & 0.424 ($\uparrow$0.9\%) & 0.405 ($\uparrow$1.2\%) & 0.418 ($\uparrow$1.4\%) & 0.415 ($\uparrow$1.2\%) \\
 & BO & 0.414 ($\uparrow$0.7\%) & 0.377 ($\uparrow$0.1\%) & 0.405 ($\uparrow$1.6\%) & 0.399 ($\uparrow$0.8\%) & 0.424 ($\uparrow$0.9\%) & 0.401 ($\uparrow$0.3\%) & 0.420 ($\uparrow$1.8\%) & 0.415 ($\uparrow$1.0\%) \\
 & MI-TGAM & \textbf{0.426 ($\uparrow$3.7\%)} & \textbf{0.390 ($\uparrow$3.6\%)} & \textbf{0.416 ($\uparrow$4.3\%)} & \textbf{0.411 ($\uparrow$3.9\%)} & \textbf{0.429 ($\uparrow$2.1\%)} & \textbf{0.410 ($\uparrow$2.6\%)} & \textbf{0.424 ($\uparrow$2.9\%)} & \textbf{0.421 ($\uparrow$2.5\%)} \\ 
\midrule
 
\multirow{10}{*}{ECL} & Clean & 0.195 (-0.0\%) & 0.206 (-0.0\%) & 0.154 (-0.0\%) & 0.185 (-0.0\%) & 0.283 (-0.0\%) & 0.291 (-0.0\%) & 0.245 (-0.0\%) & 0.273 (-0.0\%) \\
 & FGSM & 0.202 ($\uparrow$3.9\%) & 0.212 ($\uparrow$2.6\%) & 0.162 ($\uparrow$5.1\%) & 0.192 ($\uparrow$3.8\%) & 0.292 ($\uparrow$3.3\%) & 0.298 ($\uparrow$2.4\%) & 0.256 ($\uparrow$4.3\%) & 0.282 ($\uparrow$3.3\%) \\
 & BIM & 0.199 ($\uparrow$2.1\%) & 0.209 ($\uparrow$1.5\%) & 0.158 ($\uparrow$2.8\%) & 0.189 ($\uparrow$2.1\%) & 0.288 ($\uparrow$1.9\%) & 0.295 ($\uparrow$1.4\%) & 0.251 ($\uparrow$2.4\%) & 0.278 ($\uparrow$1.9\%) \\
 & PGD & 0.198 ($\uparrow$1.7\%) & 0.208 ($\uparrow$1.1\%) & 0.158 ($\uparrow$2.2\%) & 0.188 ($\uparrow$1.6\%) & 0.287 ($\uparrow$1.5\%) & 0.294 ($\uparrow$1.0\%) & 0.250 ($\uparrow$1.8\%) & 0.277 ($\uparrow$1.4\%) \\
 & MI-FGSM & 0.200 ($\uparrow$2.9\%) & 0.210 ($\uparrow$2.0\%) & 0.160 ($\uparrow$3.8\%) & 0.190 ($\uparrow$2.8\%) & 0.290 ($\uparrow$2.4\%) & 0.296 ($\uparrow$1.8\%) & 0.253 ($\uparrow$3.2\%) & 0.280 ($\uparrow$2.4\%) \\
 & ATSG & 0.201 ($\uparrow$3.3\%) & 0.211 ($\uparrow$2.3\%) & 0.161 ($\uparrow$4.3\%) & 0.191 ($\uparrow$3.2\%) & 0.293 ($\uparrow$3.5\%) & 0.299 ($\uparrow$2.7\%) & 0.257 ($\uparrow$4.5\%) & 0.283 ($\uparrow$3.5\%) \\
 & ADJM & 0.197 ($\uparrow$1.3\%) & 0.208 ($\uparrow$0.9\%) & 0.157 ($\uparrow$1.6\%) & 0.187 ($\uparrow$1.2\%) & 0.287 ($\uparrow$1.5\%) & 0.294 ($\uparrow$1.0\%) & 0.250 ($\uparrow$1.8\%) & 0.277 ($\uparrow$1.4\%) \\
 & TCA & 0.198 ($\uparrow$1.8\%) & 0.209 ($\uparrow$1.4\%) & 0.158 ($\uparrow$2.3\%) & 0.188 ($\uparrow$1.8\%) & 0.289 ($\uparrow$1.9\%) & 0.295 ($\uparrow$1.5\%) & 0.251 ($\uparrow$2.4\%) & 0.278 ($\uparrow$1.9\%) \\
 & BO & 0.196 ($\uparrow$0.6\%) & 0.207 ($\uparrow$0.3\%) & 0.155 ($\uparrow$0.7\%) & 0.186 ($\uparrow$0.5\%) & 0.285 ($\uparrow$0.7\%) & 0.292 ($\uparrow$0.5\%) & 0.247 ($\uparrow$0.8\%) & 0.275 ($\uparrow$0.6\%) \\
 & MI-TGAM & \textbf{0.205 ($\uparrow$5.4\%)} & \textbf{0.215 ($\uparrow$4.2\%)} & \textbf{0.166 ($\uparrow$7.6\%)} & \textbf{0.195 ($\uparrow$5.6\%)} & \textbf{0.297 ($\uparrow$4.9\%)} & \textbf{0.303 ($\uparrow$4.2\%)} & \textbf{0.262 ($\uparrow$6.8\%)} & \textbf{0.287 ($\uparrow$5.2\%)} \\
\midrule
 
\multirow{10}{*}{Traffic} & Clean & 0.564 (-0.0\%) & 0.694 (-0.0\%) & 0.412 (-0.0\%) & 0.557 (-0.0\%) & 0.378 (-0.0\%) & 0.353 (-0.0\%) & 0.283 (-0.0\%) & 0.338 (-0.0\%) \\
 & FGSM & 0.570 ($\uparrow$0.9\%) & 0.672 ($\downarrow$3.1\%) & 0.417 ($\uparrow$1.2\%) & 0.553 ($\downarrow$0.7\%) & 0.382 ($\uparrow$1.1\%) & 0.339 ($\downarrow$3.8\%) & 0.287 ($\uparrow$1.6\%) & 0.336 ($\downarrow$0.5\%) \\
 & BIM & 0.570 ($\uparrow$0.9\%) & 0.672 ($\downarrow$3.1\%) & 0.417 ($\uparrow$1.2\%) & 0.553 ($\downarrow$0.7\%) & 0.382 ($\uparrow$1.1\%) & 0.339 ($\downarrow$3.8\%) & 0.287 ($\uparrow$1.6\%) & 0.336 ($\downarrow$0.5\%) \\
 & PGD & 0.568 ($\uparrow$0.7\%) & 0.670 ($\downarrow$3.4\%) & 0.415 ($\uparrow$0.9\%) & 0.551 ($\downarrow$1.0\%) & 0.380 ($\uparrow$0.8\%) & 0.338 ($\downarrow$4.2\%) & 0.286 ($\uparrow$1.1\%) & 0.335 ($\downarrow$0.9\%) \\
 & MI-FGSM & 0.570 ($\uparrow$0.9\%) & 0.672 ($\downarrow$3.1\%) & 0.417 ($\uparrow$1.2\%) & 0.553 ($\downarrow$0.7\%) & 0.382 ($\uparrow$1.1\%) & 0.339 ($\downarrow$3.8\%) & 0.287 ($\uparrow$1.6\%) & 0.336 ($\downarrow$0.5\%) \\
 & ATSG & 0.569 ($\uparrow$0.7\%) & 0.672 ($\downarrow$3.1\%) & 0.416 ($\uparrow$1.0\%) & 0.552 ($\downarrow$0.8\%) & 0.382 ($\uparrow$1.1\%) & 0.339 ($\downarrow$3.8\%) & 0.287 ($\uparrow$1.6\%) & 0.336 ($\downarrow$0.4\%) \\
 & ADJM & 0.566 ($\uparrow$0.4\%) & 0.670 ($\downarrow$3.5\%) & 0.414 ($\uparrow$0.5\%) & 0.550 ($\downarrow$1.2\%) & 0.381 ($\uparrow$0.8\%) & 0.339 ($\downarrow$3.9\%) & 0.286 ($\uparrow$1.2\%) & 0.335 ($\downarrow$0.7\%) \\
 & TCA & 0.569 ($\uparrow$0.7\%) & 0.672 ($\downarrow$3.1\%) & 0.416 ($\uparrow$1.0\%) & 0.552 ($\downarrow$0.8\%) & 0.382 ($\uparrow$1.1\%) & 0.339 ($\downarrow$3.8\%) & 0.287 ($\uparrow$1.6\%) & 0.336 ($\downarrow$0.4\%) \\
 & BO & 0.565 ($\uparrow$0.2\%) & 0.668 ($\downarrow$3.8\%) & 0.412 ($\uparrow$0.2\%) & 0.549 ($\downarrow$1.5\%) & 0.379 ($\uparrow$0.3\%) & 0.337 ($\downarrow$4.5\%) & 0.284 ($\uparrow$0.6\%) & 0.333 ($\downarrow$1.3\%) \\
 & MI-TGAM & \textbf{0.577 ($\uparrow$2.2\%)} & \textbf{0.675 ($\downarrow$2.7\%)} & \textbf{0.425 ($\uparrow$3.1\%)} & \textbf{0.559 ($\uparrow$0.4\%)} & \textbf{0.387 ($\uparrow$2.6\%)} & \textbf{0.345 ($\downarrow$2.3\%)} & \textbf{0.295 ($\uparrow$4.4\%)} & \textbf{0.342 ($\uparrow$1.4\%)} \\
\bottomrule

\end{tabularx}
}
\end{table}

% \paragraph{Transfer results from Transformer-based iTransformer.}
% Finally, \cref{tab:transfer-iTransformer} illustrates the transfer effectiveness when iTransformer serves as the surrogate. 
% Despite the complex attention mechanisms inherent in Transformers, MI-TGAM successfully identifies transferable weaknesses. 
% On the Traffic dataset, MI-TGAM achieves an average MSE of 0.633, compared to 0.623 for FGSM and BIM. 
% In the ECL dataset, MI-TGAM leads with an average MSE of 0.229 against FGSM's 0.224. 
% Even when faced with the complex attention mechanisms of iTransformer, MI-TGAM robustly generates cross-model universal vulnerabilities, proving that MI-TGAM maintains a high transferability even utilizing state-of-the-art Transformer architectures as surrogates.

\begin{table}[!t]
\centering
\caption{Transfer attack results of MI-TGAM and baseline methods using iTransformer as the surrogate model.}
\label{tab:transfer-iTransformer}
\scriptsize
\resizebox{\textwidth}{!}{
    % 请在导言区添加：
% \usepackage{booktabs}
% \usepackage{multirow}

\newcolumntype{L}{>{\raggedright\arraybackslash}X} % 左对齐
\newcolumntype{C}{>{\centering\arraybackslash}X} % 居中对齐
\newcolumntype{R}{>{\raggedleft\arraybackslash}X} % 右对齐

\begin{tabularx}{1.2\textwidth}{llCCCCCCCC}
\toprule
\multirow{3}{*}{\textbf{Dataset}} & \multirow{3}{*}{\textbf{Method}} & \multicolumn{8}{c}{\textbf{Target Model (Using iTransformer as Surrogate Model)}} \\ \cmidrule(lr){3-10}
 &  & \multicolumn{4}{c}{MSE (with Degradation Percentage)} & \multicolumn{4}{c}{MAE (with Degradation Percentage)} \\ \cmidrule(lr){3-6} \cmidrule(lr){7-10}
 &  & FreTS & SegRNN & TimesNet & Average & FreTS & SegRNN & TimesNet & Average \\ \midrule
 
\multirow{10}{*}{ETT} & Clean & 0.411 (-0.0\%) & 0.377 (-0.0\%) & 0.446 (-0.0\%) & 0.411 (-0.0\%) & 0.420 (-0.0\%) & 0.400 (-0.0\%) & 0.452 (-0.0\%) & 0.424 (-0.0\%) \\
 & FGSM & 0.426 ($\uparrow$3.7\%) & 0.389 ($\uparrow$3.4\%) & 0.451 ($\uparrow$1.1\%) & 0.422 ($\uparrow$2.6\%) & 0.428 ($\uparrow$2.1\%) & 0.408 ($\uparrow$2.1\%) & 0.454 ($\uparrow$0.4\%) & 0.430 ($\uparrow$1.5\%) \\
 & BIM & 0.425 ($\uparrow$3.6\%) & 0.389 ($\uparrow$3.3\%) & 0.451 ($\uparrow$1.0\%) & 0.422 ($\uparrow$2.6\%) & 0.428 ($\uparrow$1.9\%) & 0.408 ($\uparrow$2.0\%) & 0.454 ($\uparrow$0.3\%) & 0.430 ($\uparrow$1.4\%) \\
 & PGD & 0.422 ($\uparrow$2.6\%) & 0.386 ($\uparrow$2.5\%) & 0.450 ($\uparrow$0.7\%) & 0.419 ($\uparrow$1.9\%) & 0.426 ($\uparrow$1.4\%) & 0.406 ($\uparrow$1.4\%) & 0.453 ($\uparrow$0.2\%) & 0.428 ($\uparrow$1.0\%) \\
 & MI-FGSM & 0.426 ($\uparrow$3.7\%) & 0.390 ($\uparrow$3.5\%) & 0.451 ($\uparrow$1.1\%) & 0.422 ($\uparrow$2.7\%) & 0.428 ($\uparrow$2.1\%) & 0.408 ($\uparrow$2.1\%) & 0.454 ($\uparrow$0.4\%) & 0.430 ($\uparrow$1.5\%) \\
 & ATSG & 0.424 ($\uparrow$3.2\%) & 0.388 ($\uparrow$3.1\%) & 0.450 ($\uparrow$0.9\%) & 0.421 ($\uparrow$2.4\%) & 0.429 ($\uparrow$2.1\%) & 0.409 ($\uparrow$2.4\%) & 0.453 ($\uparrow$0.3\%) & 0.431 ($\uparrow$1.5\%) \\
 & ADJM & 0.411 ($\uparrow$0.2\%) & 0.378 ($\uparrow$0.3\%) & 0.447 ($\uparrow$0.1\%) & 0.412 ($\uparrow$0.2\%) & 0.420 ($\uparrow$0.0\%) & 0.401 ($\uparrow$0.4\%) & 0.452 ($\uparrow$0.0\%) & 0.425 ($\uparrow$0.1\%) \\
 & TCA & 0.423 ($\uparrow$2.9\%) & 0.388 ($\uparrow$3.0\%) & 0.450 ($\uparrow$0.9\%) & 0.420 ($\uparrow$2.2\%) & 0.428 ($\uparrow$2.0\%) & 0.409 ($\uparrow$2.3\%) & 0.454 ($\uparrow$0.3\%) & 0.430 ($\uparrow$1.5\%) \\
 & BO & 0.414 ($\uparrow$0.7\%) & 0.377 ($\uparrow$0.1\%) & 0.450 ($\uparrow$0.8\%) & 0.414 ($\uparrow$0.6\%) & 0.424 ($\uparrow$0.9\%) & 0.401 ($\uparrow$0.3\%) & \textbf{0.457 ($\uparrow$1.1\%)} & 0.427 ($\uparrow$0.8\%) \\
 & MI-TGAM & \textbf{0.439 ($\uparrow$6.8\%)} & \textbf{0.399 ($\uparrow$5.8\%)} & \textbf{0.453 ($\uparrow$1.5\%)} & \textbf{0.430 ($\uparrow$4.6\%)} & \textbf{0.438 ($\uparrow$4.3\%)} & \textbf{0.416 ($\uparrow$4.0\%)} & 0.455 ($\uparrow$0.6\%) & \textbf{0.436 ($\uparrow$2.9\%)} \\
\midrule
 
\multirow{10}{*}{ECL} & Clean & 0.195 (-0.0\%) & 0.206 (-0.0\%) & 0.259 (-0.0\%) & 0.220 (-0.0\%) & 0.283 (-0.0\%) & 0.291 (-0.0\%) & 0.341 (-0.0\%) & 0.305 (-0.0\%) \\
 & FGSM & 0.203 ($\uparrow$4.0\%) & 0.212 ($\uparrow$2.7\%) & 0.256 ($\downarrow$1.0\%) & 0.224 ($\uparrow$1.6\%) & 0.293 ($\uparrow$3.5\%) & 0.298 ($\uparrow$2.5\%) & 0.339 ($\downarrow$0.8\%) & 0.310 ($\uparrow$1.6\%) \\
 & BIM & 0.199 ($\uparrow$2.1\%) & 0.209 ($\uparrow$1.3\%) & 0.256 ($\downarrow$1.0\%) & 0.221 ($\uparrow$0.6\%) & 0.288 ($\uparrow$1.7\%) & 0.294 ($\uparrow$1.1\%) & 0.338 ($\downarrow$1.0\%) & 0.307 ($\uparrow$0.5\%) \\
 & PGD & 0.199 ($\uparrow$2.1\%) & 0.209 ($\uparrow$1.2\%) & 0.256 ($\downarrow$1.1\%) & 0.221 ($\uparrow$0.6\%) & 0.288 ($\uparrow$1.7\%) & 0.294 ($\uparrow$1.1\%) & 0.338 ($\downarrow$1.0\%) & 0.307 ($\uparrow$0.5\%) \\
 & MI-FGSM & 0.202 ($\uparrow$3.5\%) & 0.211 ($\uparrow$2.2\%) & 0.256 ($\downarrow$1.0\%) & 0.223 ($\uparrow$1.3\%) & 0.291 ($\uparrow$2.9\%) & 0.297 ($\uparrow$2.0\%) & 0.338 ($\downarrow$0.9\%) & 0.309 ($\uparrow$1.2\%) \\
 & ATSG & 0.202 ($\uparrow$3.8\%) & 0.212 ($\uparrow$2.6\%) & 0.256 ($\downarrow$1.0\%) & 0.223 ($\uparrow$1.6\%) & 0.294 ($\uparrow$4.0\%) & 0.299 ($\uparrow$3.0\%) & 0.339 ($\downarrow$0.7\%) & 0.311 ($\uparrow$1.9\%) \\
 & ADJM & 0.198 ($\uparrow$1.4\%) & 0.208 ($\uparrow$0.9\%) & 0.256 ($\downarrow$1.1\%) & 0.221 ($\uparrow$0.3\%) & 0.288 ($\uparrow$1.6\%) & 0.294 ($\uparrow$1.1\%) & 0.338 ($\downarrow$0.9\%) & 0.307 ($\uparrow$0.5\%) \\
 & TCA & 0.198 ($\uparrow$1.8\%) & 0.209 ($\uparrow$1.2\%) & 0.256 ($\downarrow$1.1\%) & 0.221 ($\uparrow$0.5\%) & 0.288 ($\uparrow$1.8\%) & 0.294 ($\uparrow$1.3\%) & 0.338 ($\downarrow$1.0\%) & 0.307 ($\uparrow$0.6\%) \\
 & BO & 0.196 ($\uparrow$0.6\%) & 0.207 ($\uparrow$0.3\%) & 0.256 ($\downarrow$1.0\%) & 0.220 ($\downarrow$0.1\%) & 0.285 ($\uparrow$0.7\%) & 0.292 ($\uparrow$0.5\%) & 0.338 ($\downarrow$0.9\%) & 0.305 ($\uparrow$0.0\%) \\
 & MI-TGAM & \textbf{0.210 ($\uparrow$7.8\%)} & \textbf{0.218 ($\uparrow$5.6\%)} & \textbf{0.259 ($\uparrow$0.1\%)} & \textbf{0.229 ($\uparrow$4.1\%)} & \textbf{0.301 ($\uparrow$6.4\%)} & \textbf{0.305 ($\uparrow$5.0\%)} & \textbf{0.342 ($\uparrow$0.2\%)} & \textbf{0.316 ($\uparrow$3.6\%)} \\
\midrule
 
\multirow{10}{*}{Traffic} & Clean & 0.564 (-0.0\%) & 0.694 (-0.0\%) & 0.611 (-0.0\%) & 0.623 (-0.0\%) & 0.378 (-0.0\%) & 0.353 (-0.0\%) & 0.333 (-0.0\%) & 0.354 (-0.0\%) \\
 & FGSM & 0.576 ($\uparrow$2.0\%) & 0.675 ($\downarrow$2.7\%) & 0.617 ($\uparrow$1.1\%) & 0.623 ($\downarrow$0.0\%) & 0.385 ($\uparrow$2.0\%) & 0.342 ($\downarrow$3.0\%) & 0.337 ($\uparrow$1.4\%) & 0.355 ($\uparrow$0.2\%) \\
 & BIM & 0.576 ($\uparrow$2.0\%) & 0.675 ($\downarrow$2.7\%) & 0.617 ($\uparrow$1.1\%) & 0.623 ($\downarrow$0.0\%) & 0.385 ($\uparrow$2.0\%) & 0.342 ($\downarrow$3.0\%) & 0.337 ($\uparrow$1.4\%) & 0.355 ($\uparrow$0.2\%) \\
 & PGD & 0.573 ($\uparrow$1.5\%) & 0.673 ($\downarrow$3.0\%) & 0.616 ($\uparrow$0.9\%) & 0.621 ($\downarrow$0.4\%) & 0.383 ($\uparrow$1.4\%) & 0.340 ($\downarrow$3.6\%) & 0.337 ($\uparrow$1.1\%) & 0.353 ($\downarrow$0.3\%) \\
 & MI-FGSM & 0.576 ($\uparrow$2.0\%) & 0.675 ($\downarrow$2.7\%) & 0.617 ($\uparrow$1.1\%) & 0.623 ($\downarrow$0.0\%) & 0.385 ($\uparrow$2.0\%) & 0.342 ($\downarrow$3.0\%) & 0.337 ($\uparrow$1.4\%) & 0.355 ($\uparrow$0.2\%) \\
 & ATSG & 0.574 ($\uparrow$1.7\%) & 0.675 ($\downarrow$2.7\%) & 0.617 ($\uparrow$1.1\%) & 0.622 ($\downarrow$0.1\%) & 0.387 ($\uparrow$2.5\%) & 0.344 ($\downarrow$2.4\%) & 0.338 ($\uparrow$1.5\%) & 0.356 ($\uparrow$0.6\%) \\
 & ADJM & 0.567 ($\uparrow$0.4\%) & 0.670 ($\downarrow$3.4\%) & 0.612 ($\uparrow$0.3\%) & 0.616 ($\downarrow$1.1\%) & 0.381 ($\uparrow$0.8\%) & 0.339 ($\downarrow$4.0\%) & 0.335 ($\uparrow$0.6\%) & 0.351 ($\downarrow$0.8\%) \\
 & TCA & 0.574 ($\uparrow$1.7\%) & 0.675 ($\downarrow$2.7\%) & 0.617 ($\uparrow$1.1\%) & 0.622 ($\downarrow$0.1\%) & 0.387 ($\uparrow$2.5\%) & 0.344 ($\downarrow$2.4\%) & 0.338 ($\uparrow$1.5\%) & 0.356 ($\uparrow$0.6\%) \\
 & BO & 0.565 ($\uparrow$0.2\%) & 0.668 ($\downarrow$3.8\%) & 0.613 ($\uparrow$0.4\%) & 0.615 ($\downarrow$1.2\%) & 0.379 ($\uparrow$0.3\%) & 0.337 ($\downarrow$4.5\%) & 0.335 ($\uparrow$0.8\%) & 0.350 ($\downarrow$1.1\%) \\
 & MI-TGAM & \textbf{0.589 ($\uparrow$4.3\%)} & \textbf{0.685 ($\downarrow$1.3\%)} & \textbf{0.624 ($\uparrow$2.2\%)} & \textbf{0.633 ($\uparrow$1.5\%)} & \textbf{0.393 ($\uparrow$4.2\%)} & \textbf{0.350 ($\downarrow$0.7\%)} & \textbf{0.341 ($\uparrow$2.5\%)} & \textbf{0.361 ($\uparrow$2.0\%)}
\\ \bottomrule

\end{tabularx}
}
\end{table}

\begin{figure}[!t]
    \centering
    \includegraphics[width=1\linewidth]{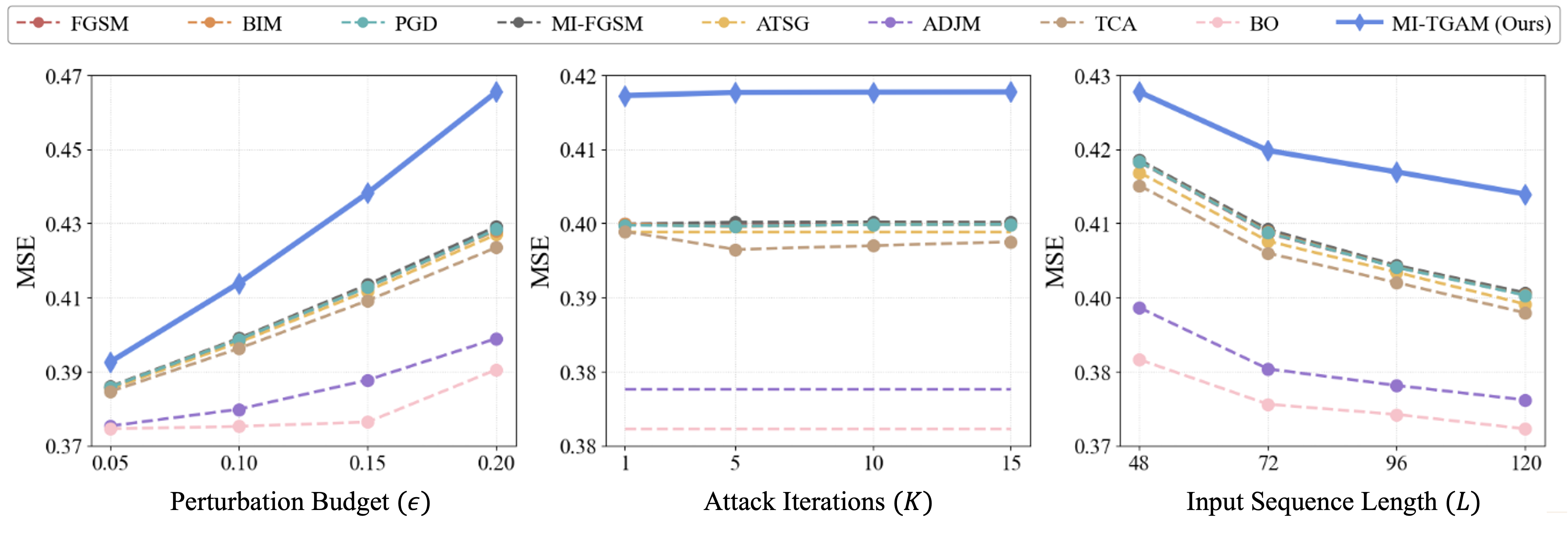}
    \caption{Sensitivity of attack performance (MSE) to key hyperparameters ($\epsilon$, $K$, and $L$) on the ETT dataset using SegRNN as the surrogate and target model.}
    \label{fig:sensitivity}
\end{figure}

\subsection{Sensitivity Analysis}

To validate the robustness and stability of MI-TGAM under varying attack parameters and forecasting task configurations, we conduct a comprehensive sensitivity analysis on three key hyperparameters: the perturbation budget $\epsilon$, the number of attack iterations $K$, and the input sequence length $L$ (while keeping the forecast horizon $H=96$ fixed). 
We perform this analysis on the ETT dataset using SegRNN as both the surrogate and target model (white-box setting), with results shown in \cref{fig:sensitivity}.

We vary the perturbation budget $\epsilon$ over the set $\{0.05, 0.10, 0.15, 0.20\}$, corresponding to 5\%, 10\%, 15\%, and 20\% of the original magnitude. 
The results show a clear increasing trend where larger budgets lead to higher MSE degradation for all attack methods, as expected. 
Notably, the performance gap between MI-TGAM and the strongest baselines (PGD and MI-FGSM) widens as $\epsilon$ increases, demonstrating that MI-TGAM can more effectively exploit larger perturbation budgets to maximize attack effectiveness.

We examine the sensitivity to the number of iterations $K \in \{1, 5, 10, 15\}$. 
MI-TGAM shows consistent advantages over all baselines across all iteration settings.
Remarkably, MI-TGAM achieves near-peak attack performance even with a single iteration ($K=1$), which is equivalent to the TGSM baseline (TGAM without momentum and iteration).
This demonstrates the effectiveness of the timestamp-wise gradient accumulation mechanism, which enriches gradient information even in single-step attacks.
% As $K$ increases, MI-TGAM's performance continues to improve, with the momentum mechanism providing additional benefits for multi-step optimization. 

Finally, we evaluate the impact of varying the input sequence length $L \in \{48, 72, 96, 120\}$ while keeping the forecast horizon $H=96$ fixed. 
The results show that all methods exhibit progressively reduced attack effectiveness (lower MSE) as the input sequence length increases. 
This can be attributed to the fact that longer input sequences generally provide more contextual information, improving the model's forecasting stability and robustness, thereby increasing the difficulty of successful adversarial attacks. 
Despite this trend, MI-TGAM consistently outperforms all baseline methods across all input sequence lengths, maintaining a substantial performance advantage even when $L=120$.

In summary, the sensitivity analysis demonstrates the stable and robust performance of MI-TGAM across different hyperparameter settings and task configurations.
These results confirm that MI-TGAM is not only effective under the default settings but also maintains its superiority across a wide range of practical scenarios, making it a reliable and versatile attack method for time series forecasting models.

\begin{figure}[!t]
    \centering
    \begin{subfigure}[b]{\linewidth}
        \centering
        \includegraphics[width=\linewidth]{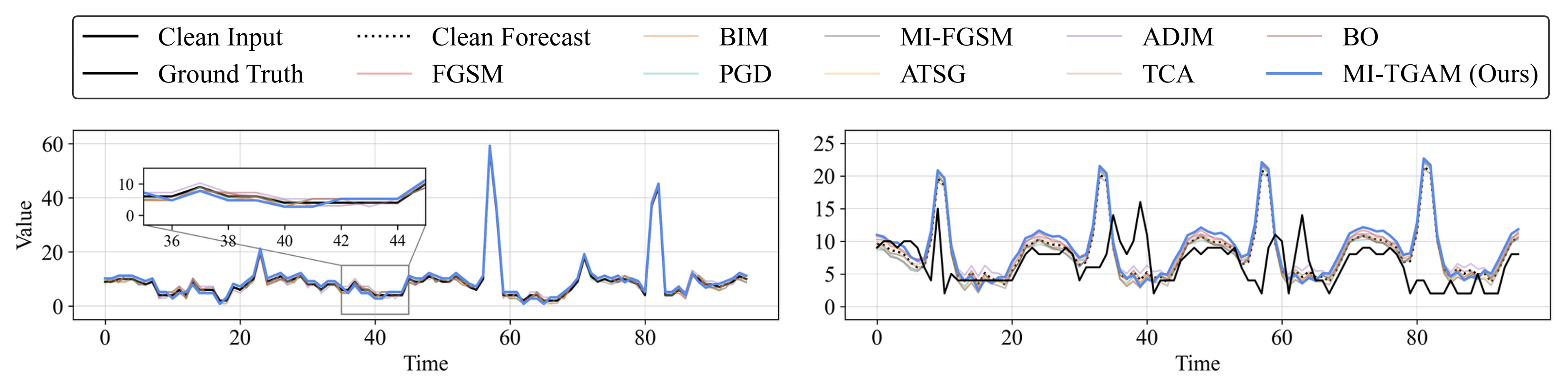}
        \vspace{-2em}
        \caption{$\epsilon=0.05$}
    \end{subfigure}

    \begin{subfigure}[b]{\linewidth}
        \centering
        \includegraphics[width=\linewidth]{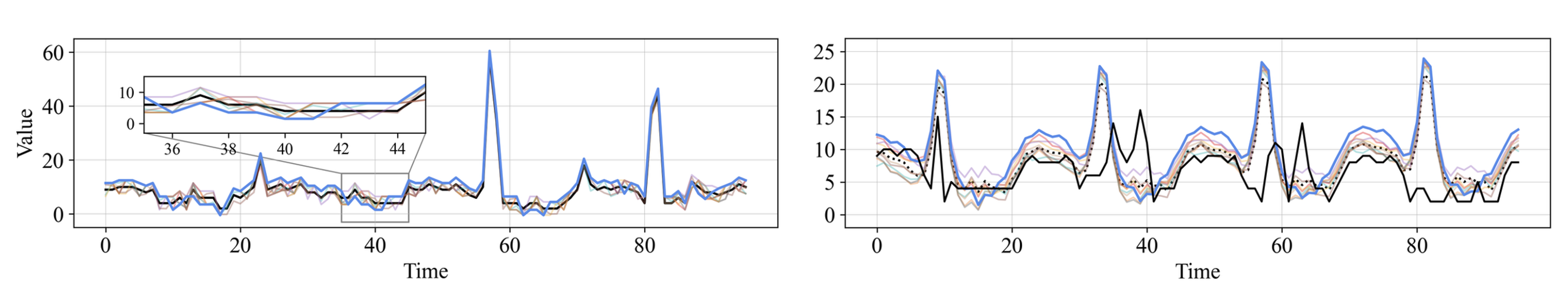}
        \vspace{-2em}
        \caption{$\epsilon=0.1$}
    \end{subfigure}

    \begin{subfigure}[b]{\linewidth}
        \centering
        \includegraphics[width=\linewidth]{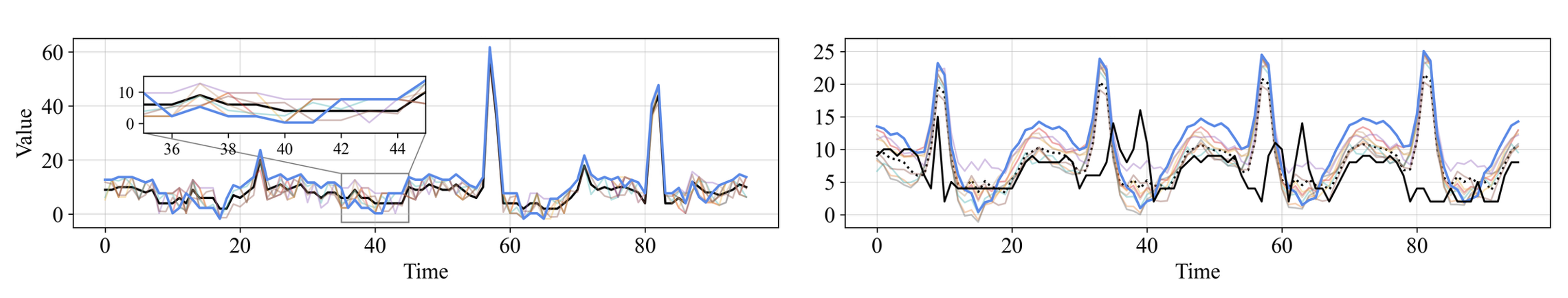}
        \vspace{-2em}
        \caption{$\epsilon=0.15$}
    \end{subfigure}

    \begin{subfigure}[b]{\linewidth}
        \centering
        \includegraphics[width=\linewidth]{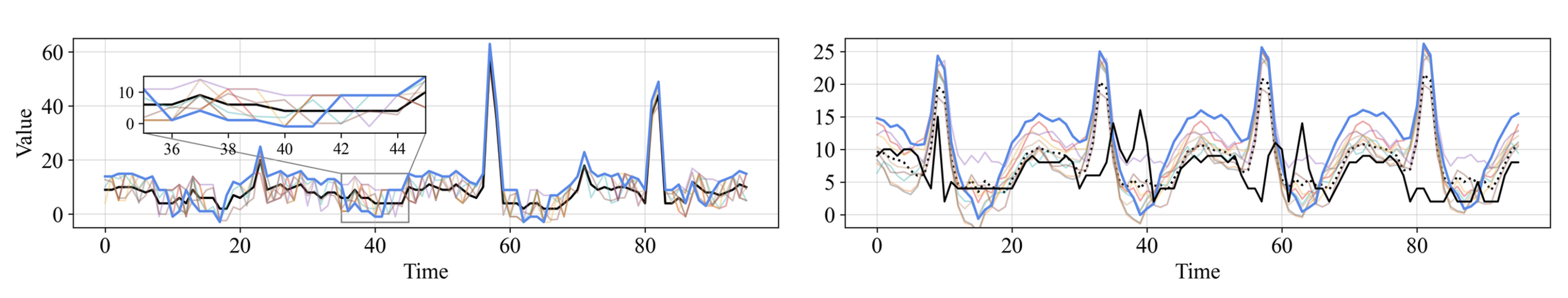}
        \vspace{-2em}
        \caption{$\epsilon=0.2$}
    \end{subfigure}
    \caption{Visualization of the adversary conducted by all methods using SegRNN as the surrogate and the target model on ECL dataset with varying perturbation budgets $\epsilon$. The left panels show the clean and perturbed input samples, the right panels show the corresponding forecasting outputs.}
    \label{fig:visual-epsilon}
\end{figure}

\subsection{Visual Analysis}

To intuitively understand the effectiveness of MI-TGAM, we visualize adversarial examples and the corresponding model forecasts in \cref{fig:visual-epsilon}. 
These visualizations are derived from a randomly selected sample and a specific variable on the Electricity dataset, comparing MI-TGAM against baseline methods using SegRNN as both the surrogate and target model (white-box setting), with perturbation budgets $\epsilon \in \{0.05, 0.10, 0.15, 0.20\}$.

From the left subfigures of \cref{fig:visual-epsilon}, we observe that despite the increasing perturbation budget, the adversarial examples generated by MI-TGAM maintain high visual similarity to the original input in terms of overall trend and temporal patterns. 
Unlike baseline methods that often introduce high-frequency, chaotic noise that is easily detectable, MI-TGAM produces perturbations with a more regular and smooth pattern that preserves the natural characteristics of time series data.
This makes MI-TGAM's perturbations more stealthy and harder to detect, which is crucial for practical adversarial attacks.

As shown in the right subfigures of \cref{fig:visual-epsilon}, the forecasting results demonstrate that MI-TGAM does not merely introduce random variance in the predictions like other baseline methods.
Instead, MI-TGAM induces significant deviations from the ground truth. 
Specifically, we observe that MI-TGAM amplifies prediction errors in critical regions such as peaks, valleys, and high-volatility areas, where accurate forecasting is most important.
This targeted error amplification directly enhances the adversarial effect, making MI-TGAM particularly effective at degrading the practical utility of forecasting models.
In contrast, baseline methods produce more scattered errors that are less impactful on the overall forecasting quality.

Overall, these visualizations confirm that MI-TGAM generates adversarial examples that are both effective (causing significant prediction errors) and stealthy (maintaining natural appearance), making it a powerful and practical attack method for time series forecasting models.

\subsection{Comparison in Temporal Inconsistent Settings}

To further verify attack performance of proposed MI-TGAM, we compare it with all baseline methods in temporal inconsistent settings. 
Unlike the former settings, here we relax the TUAP constraint, allowing the same time stamp across different samples could have different perturbations. 
The experiment is conducted on the Traffic dataset in both white-box attack and transfer attack scenarios using TimesNet as the surrogate model.
The results are shown in \cref{tab:time-inconsistent_comparison}.

\begin{table}[!t]
\centering
\scriptsize
\caption{The white-box attack and transfer attack results of MI-TGAM and baseline methods using TimesNet as the surrogate model on Traffic dataset in temporal inconsistent setting. White-box attack results are presented in \colorbox[HTML]{EFEFEF}{gray}.}
\label{tab:time-inconsistent_comparison}
\newcolumntype{L}{>{\raggedright\arraybackslash}X} 
\newcolumntype{C}{>{\centering\arraybackslash}X} 

\begin{tabularx}{\textwidth}{LCCCCC}
\toprule
\multirow{2}{*}{\textbf{Method}} & \multicolumn{5}{c}{Degradation Percentage on MSE} \\ \cmidrule(lr){2-6} 
 & TimesNet & FreTS & SegRNN & iTransformer & Average \\ \midrule
FGSM & \cellcolor[HTML]{EFEFEF}85.88\% & 1.49\% & -3.03\% & 2.04\% & 0.17\% \\
BIM & \cellcolor[HTML]{EFEFEF}85.88\% & 1.49\% & -3.03\% & 2.04\% & 0.17\% \\
PGD & \cellcolor[HTML]{EFEFEF}59.57\% & 1.06\% & -3.31\% & 1.47\% & -0.26\% \\
MI-FGSM & \cellcolor[HTML]{EFEFEF}85.88\% & 1.49\% & -3.03\% & 2.04\% & 0.17\% \\
ATSG & \cellcolor[HTML]{EFEFEF}103.66\% & 1.09\% & -3.03\% & 1.53\% & -0.14\% \\
ADJM & \cellcolor[HTML]{EFEFEF}\textbf{105.25\%} & 0.31\% & -3.49\% & 0.52\% & -0.89\% \\
TCA & \cellcolor[HTML]{EFEFEF}103.66\% & 1.09\% & -3.03\% & 1.53\% & -0.14\% \\
BO & \cellcolor[HTML]{EFEFEF}0.38\% & 0.28\% & -3.70\% & 0.42\% & -1.00\% \\
MI-TGAM & \cellcolor[HTML]{EFEFEF}15.48\% & \textbf{2.18\%} & \textbf{-2.69\%} & \textbf{3.12\%} & \textbf{0.87\%}
\\
\bottomrule
\end{tabularx} % Ensure this file contains the data from the provided image
\end{table}

% \paragraph{White-box Attack Performance.} 
In the the white-box attack, the attack strength of MI-TGAM is lower than that of unconstrained baseline methods, which is as expected. This disparity arises because temporal inconsistent methods possess a significantly larger optimization space, allowing them to exploit per-window gradients independently without adhering to temporal constraints. 
The results empirically show that imposing a strict temporal-consistency constraint necessitates a trade-off in local optimization potency.

% \paragraph{Transfer Attack Performance.} 
Despite the deficit in white-box attack, MI-TGAM demonstrates superior performance in cross-model transfer attacks, consistently achieving the highest transfer attack results.
These findings suggest that, compared to unconstrained baseline methods, MI-TGAM can enhance the attack strength in transfer attacks under temporal inconsistent settings.

\begin{figure}[!t]
    \centering
    \includegraphics[width=1\linewidth]{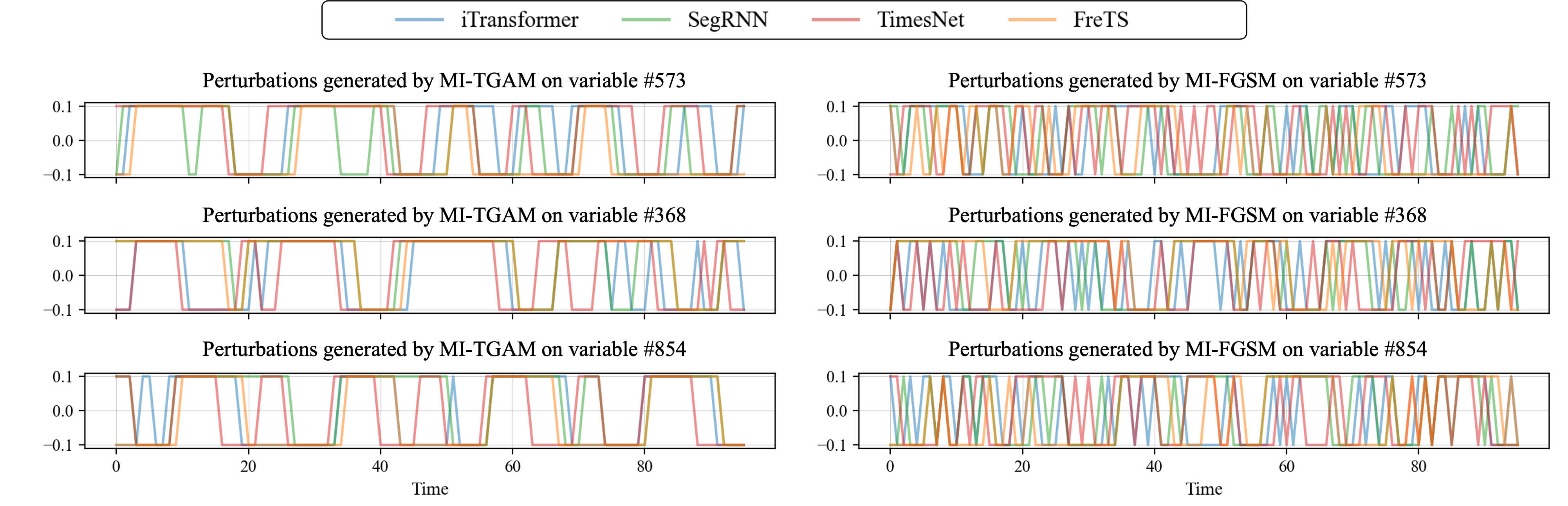}
    \caption{Perturbations generated by MI-TGAM (left) and MI-FGSM (right) under four surrogate models with three randomly selected variable within a randomly selected sample on Traffic dataset.}
    \label{fig:visual_comparison}
\end{figure}

To elucidate the underlying mechanism driving the high transferability of MI-TGAM, we visualize the perturbations generated by MI-TGAM and MI-FGSM across three randomly selected variable within a randomly selected sample on Traffic dataset in \cref{fig:visual_comparison}.

% \subsection{Intrinsics of Superior Transferability}

% \paragraph{Cross-Model Perturbation Stability.} 
Comparing the perturbations on different surrogate models, the perturbations generated by MI-TGAM (left column) exhibit a high degree of consistency in the perturbation trajectory.
In contrast, the perturbations generated by MI-FGSM (right column) are chaotic and divergent, with significant differences between the different surrogate models. 
These results indicate that MI-TGAM successfully extracts robust vulnerability patterns that are shared across different neural architectures.

\section{Conclusion}

This paper aims to address the critical yet overlooked issue of temporal inconsistency in adversarial attacks on time series forecasting. 
% Existing methods optimize perturbations for individual sliding windows independently, resulting in conflicting values for the same timestamp and rendering the attacks unrealizable in real-world scenarios. 
To this end, we introduce the concept of Temporally Unified Adversarial Perturbations (TUAPs), propose the Timestamp-wise Gradient Accumulation Method (TGAM), and further incorporate TGAM into a momentum-based iterative framework (MI-TGAM). 
% By aggregating local gradient information across the overlapping samples, our method generates the temporally unified perturbations that maintains temporal consistency while effectively degrading model performance.
Comprehensive experiments on benchmark datasets and diverse model architectures demonstrate the effectiveness and superiority of our proposed method. 
In the future, we will extend TUAPs into attacking time series classification models and establishing a robust defense mechanism against such temporally consistent attacks.

% opens promising avenues for developing robust defense mechanisms against physically consistent attacks and extending the temporal unification principles to other time-sensitive tasks and modalities.
% In white-box scenarios, MI-TGAM significantly outperforms traditional window-wise baselines by effectively exploiting global gradient information. 
% More importantly, we identify a counter-intuitive phenomenon: the strict physical consistency constraint serves as a powerful regularizer that enhances cross-model transferability. 
% By filtering out model-specific local noise, MI-TGAM captures intrinsic vulnerability patterns inherent in time series data, allowing the adversarial signal to generalize effectively across different neural structures.

% Looking ahead, our work opens several promising research avenues. 
% First, the MI-TGAM framework provides a rigorous baseline for developing more robust defense mechanisms, such as physically-grounded adversarial training. 
% Second, the principles of temporal unification can be extended beyond forecasting to other time-sensitive tasks, including time series classification and multi-modal domains like audio and action recognition. 
% Since sliding-window processing is a fundamental paradigm in these fields, the physical consistency constraint proposed in this work has broad implications for designing physically realizable and cross-modal adversarial attacks.

\bibliographystyle{unsrtnat}
\bibliography{references}  %%% Remove comment to use the external .bib file (using bibtex).

@article{aslam_survey_2021,
	title = {A survey on deep learning methods for power load and renewable energy forecasting in smart microgrids},
	volume = {144},
	issn = {1364-0321},
	url = {https://www.sciencedirect.com/science/article/pii/S1364032121002847},
	doi = {10.1016/j.rser.2021.110992},
	pages = {110992},
	journal = {Renewable and Sustainable Energy Reviews},
	author = {Aslam, Sheraz and Herodotou, Herodotos and Mohsin, Syed Muhammad and Javaid, Nadeem and Ashraf, Nouman and Aslam, Shahzad},
	year = {2021},
}

@article{olorunnimbe_deep_2023,
	title = {Deep learning in the stock market—a systematic survey of practice, backtesting, and applications},
	volume = {56},
	issn = {1573-7462},
	url = {https://doi.org/10.1007/s10462-022-10226-0},
	doi = {10.1007/s10462-022-10226-0},
	pages = {2057--2109},
	number = {3},
	journal = {Artificial Intelligence Review},
	author = {Olorunnimbe, Kenniy and Viktor, Herna},
	year = {2023},
}

@inproceedings{liu_itransformer_2023,
	title = {{iTransformer}: Inverted Transformers Are Effective for Time Series Forecasting},
	url = {https://openreview.net/forum?id=JePfAI8fah},
	booktitle = {The Twelfth International Conference on Learning Representations},
	author = {Liu, Yong and Hu, Tengge and Zhang, Haoran and Wu, Haixu and Wang, Shiyu and Ma, Lintao and Long, Mingsheng},
	year = {2023},
}

@inproceedings{wu_timesnet_2022,
	title = {{TimesNet}: Temporal 2D-Variation Modeling for General Time Series Analysis},
	url = {https://openreview.net/forum?id=ju_Uqw384Oq},
	booktitle = {The Eleventh International Conference on Learning Representations},
	author = {Wu, Haixu and Hu, Tengge and Liu, Yong and Zhou, Hang and Wang, Jianmin and Long, Mingsheng},
	year = {2022},
}

@misc{szegedy_intriguing_2014,
	title = {Intriguing properties of neural networks},
	url = {http://arxiv.org/abs/1312.6199},
	doi = {10.48550/arXiv.1312.6199},
	author = {Szegedy, Christian and Zaremba, Wojciech and Sutskever, Ilya and Bruna, Joan and Erhan, Dumitru and Goodfellow, Ian and Fergus, Rob},
	year = {2014},
}

@misc{goodfellow_explaining_2015,
	title = {Explaining and {Harnessing} {Adversarial} {Examples}},
	url = {http://arxiv.org/abs/1412.6572},
	doi = {10.48550/arXiv.1412.6572},
	author = {Goodfellow, Ian J. and Shlens, Jonathon and Szegedy, Christian},
	year = {2015},
}

@misc{madry_towards_2019,
	title = {Towards {Deep} {Learning} {Models} {Resistant} to {Adversarial} {Attacks}},
	url = {http://arxiv.org/abs/1706.06083},
	doi = {10.48550/arXiv.1706.06083},
	author = {Madry, Aleksander and Makelov, Aleksandar and Schmidt, Ludwig and Tsipras, Dimitris and Vladu, Adrian},
	year = {2019},
}

@misc{dong_boosting_2018,
	title = {Boosting {Adversarial} {Attacks} with {Momentum}},
	url = {http://arxiv.org/abs/1710.06081},
	doi = {10.48550/arXiv.1710.06081},
	author = {Dong, Yinpeng and Liao, Fangzhou and Pang, Tianyu and Su, Hang and Zhu, Jun and Hu, Xiaolin and Li, Jianguo},
	year = {2018},
}

@misc{kurakin_adversarial_2017,
	title = {Adversarial examples in the physical world},
	url = {http://arxiv.org/abs/1607.02533},
	doi = {10.48550/arXiv.1607.02533},
	author = {Kurakin, Alexey and Goodfellow, Ian and Bengio, Samy},
	year = {2017},
}

@inproceedings{mode_adversarial_2020,
	title = {Adversarial {Examples} in {Deep} {Learning} for {Multivariate} {Time} {Series} {Regression}},
	url = {https://ieeexplore.ieee.org/abstract/document/9425190},
	doi = {10.1109/AIPR50011.2020.9425190},
	booktitle = {2020 {IEEE} {Applied} {Imagery} {Pattern} {Recognition} {Workshop} ({AIPR})},
	author = {Mode, Gautam Raj and Hoque, Khaza Anuarul},
	year = {2020},
	pages = {1--10},
}

@article{shen_temporal_2025,
	title = {Temporal characteristics-based adversarial attacks on time series forecasting},
	volume = {264},
	issn = {0957-4174},
	url = {https://www.sciencedirect.com/science/article/pii/S0957417424028173},
	doi = {10.1016/j.eswa.2024.125950},
	journal = {Expert Systems with Applications},
	author = {Shen, Ziyu and Li, Yun},
	year = {2025},
	pages = {125950},
}

@article{wu_small_2022,
	title = {Small perturbations are enough: {Adversarial} attacks on time series prediction},
	volume = {587},
	issn = {0020-0255},
	url = {https://www.sciencedirect.com/science/article/pii/S0020025521011178},
	doi = {10.1016/j.ins.2021.11.007},
	journal = {Information Sciences},
	author = {Wu, Tao and Wang, Xuechun and Qiao, Shaojie and Xian, Xingping and Liu, Yanbing and Zhang, Liang},
	year = {2022},
	pages = {794--812},
}

@article{jiao_gradient-based_2024,
	title = {A {Gradient}-{Based} {Wind} {Power} {Forecasting} {Attack} {Method} {Considering} {Point} and {Direction} {Selection}},
	volume = {15},
	issn = {1949-3061},
	url = {https://ieeexplore.ieee.org/abstract/document/10304287},
	doi = {10.1109/TSG.2023.3325390},
	number = {3},
	journal = {IEEE Transactions on Smart Grid},
	author = {Jiao, Runhai and Han, Zhuoting and Liu, Xuan and Zhou, Changyu and Du, Min},
	year = {2024},
	pages = {3178--3192},
}

@article{heinrich_targeted_2024,
	title = {Targeted adversarial attacks on wind power forecasts},
	volume = {113},
	issn = {1573-0565},
	url = {https://doi.org/10.1007/s10994-023-06396-9},
	doi = {10.1007/s10994-023-06396-9},
	number = {2},
	journal = {Machine Learning},
	author = {Heinrich, René and Scholz, Christoph and Vogt, Stephan and Lehna, Malte},
	year = {2024},
	pages = {863--889},
}

@article{xu_novel_2025,
	title = {A {Novel} {Adversarial} {Attack} {Method} for {Time}-{Series} {Regression} {Models} in {IIoT}-{Based} {Digital} {Twins}},
	volume = {12},
	issn = {2327-4662},
	url = {https://ieeexplore.ieee.org/abstract/document/11004038},
	doi = {10.1109/JIOT.2025.3569857},
	number = {15},
	journal = {IEEE Internet of Things Journal},
	author = {Xu, Bo and Liu, Zhiqiang and Zhu, Haolin and Dong, Bingqing and Zhao, Bo and Yan, Ben and Wei, Jun},
	year = {2025},
	pages = {29278--29290},
}

@article{belkhouja_dynamic_2023,
	title = {Dynamic {Time} {Warping} {Based} {Adversarial} {Framework} for {Time}-{Series} {Domain}},
	volume = {45},
	issn = {1939-3539},
	url = {https://ieeexplore.ieee.org/abstract/document/9970291},
	doi = {10.1109/TPAMI.2022.3224754},
	number = {6},
	journal = {IEEE Transactions on Pattern Analysis and Machine Intelligence},
	author = {Belkhouja, Taha and Yan, Yan and Doppa, Janardhan Rao},
	year = {2023},
	pages = {7353--7366},
}

@misc{lin_segrnn_2023,
	title = {{SegRNN}: {Segment} {Recurrent} {Neural} {Network} for {Long}-{Term} {Time} {Series} {Forecasting}},
	url = {http://arxiv.org/abs/2308.11200},
	doi = {10.48550/arXiv.2308.11200},
	author = {Lin, Shengsheng and Lin, Weiwei and Wu, Wentai and Zhao, Feiyu and Mo, Ruichao and Zhang, Haotong},
	year = {2023},
}

@inproceedings{wang_investigation_2023,
	title = {Investigation of {Artificial} {Intelligence} {Vulnerability} in {Smart} {Grids}: {A} {Case} from {Solar} {Energy} {Forecasting}},
	shorttitle = {Investigation of {Artificial} {Intelligence} {Vulnerability} in {Smart} {Grids}},
	url = {https://ieeexplore.ieee.org/document/10512845/},
	doi = {10.1109/EI259745.2023.10512845},
	language = {en-US},
	urldate = {2026-01-10},
	booktitle = {2023 {IEEE} 7th {Conference} on {Energy} {Internet} and {Energy} {System} {Integration} ({EI2})},
	author = {Wang, Qihan and Ruan, Jiaqi and Meng, Xiangrui and Zhu, Yifan and Liang, Gaoqi and Zhao, Junhua},
	month = dec,
	year = {2023},
	pages = {5207--5212},
}

@article{yi_frequency-domain_2023,
	title = {Frequency-domain {MLPs} are {More} {Effective} {Learners} in {Time} {Series} {Forecasting}},
	volume = {36},
	url = {https://proceedings.neurips.cc/paper_files/paper/2023/hash/f1d16af76939f476b5f040fd1398c0a3-Abstract-Conference.html},
	language = {en},
	urldate = {2026-01-19},
	journal = {Advances in Neural Information Processing Systems},
	author = {Yi, Kun and Zhang, Qi and Fan, Wei and Wang, Shoujin and Wang, Pengyang and He, Hui and An, Ning and Lian, Defu and Cao, Longbing and Niu, Zhendong},
	month = dec,
	year = {2023},
	pages = {76656--76679},
}

@inproceedings{zhangCraftingAdversarialExamples2021,
  title = {Crafting {{Adversarial Examples}} for {{Neural Machine Translation}}},
  booktitle = {Proceedings of the 59th {{Annual Meeting}} of the {{Association}} for {{Computational Linguistics}} and the 11th {{International Joint Conference}} on {{Natural Language Processing}} ({{Volume}} 1: {{Long Papers}})},
  author = {Zhang, Xinze and Zhang, Junzhe and Chen, Zhenhua and He, Kun},
  editor = {Zong, Chengqing and Xia, Fei and Li, Wenjie and Navigli, Roberto},
  year = 2021,
  month = aug,
  pages = {1967--1977},
  publisher = {Association for Computational Linguistics},
  address = {Online},
  doi = {10.18653/v1/2021.acl-long.153},
  urldate = {2026-01-29}
}

@inproceedings{renGeneratingNaturalLanguage2019,
  title = {Generating {{Natural Language Adversarial Examples}} through {{Probability Weighted Word Saliency}}},
  booktitle = {Proceedings of the 57th {{Annual Meeting}} of the {{Association}} for {{Computational Linguistics}}},
  author = {Ren, Shuhuai and Deng, Yihe and He, Kun and Che, Wanxiang},
  editor = {Korhonen, Anna and Traum, David and M{\`a}rquez, Llu{\'i}s},
  year = 2019,
  month = jul,
  pages = {1085--1097},
  publisher = {Association for Computational Linguistics},
  address = {Florence, Italy},
  doi = {10.18653/v1/P19-1103},
  urldate = {2026-01-29}
}

@article{dingBlackBoxAdversarialAttack2023,
  title = {Black-{{Box Adversarial Attack}} on {{Time Series Classification}}},
  author = {Ding, Daizong and Zhang, Mi and Feng, Fuli and Huang, Yuanmin and Jiang, Erling and Yang, Min},
  year = 2023,
  month = jun,
  journal = {Proceedings of the AAAI Conference on Artificial Intelligence},
  volume = {37},
  number = {6},
  pages = {7358--7368},
  issn = {2374-3468},
  doi = {10.1609/aaai.v37i6.25896},
  urldate = {2025-11-22},
  copyright = {Copyright (c) 2023 Association for the Advancement of Artificial Intelligence},
  langid = {english}
}

@inproceedings{modeAdversarialExamplesDeep2020,
  title = {Adversarial {{Examples}} in {{Deep Learning}} for {{Multivariate Time Series Regression}}},
  booktitle = {2020 {{IEEE Applied Imagery Pattern Recognition Workshop}} ({{AIPR}})},
  author = {Mode, Gautam Raj and Hoque, Khaza Anuarul},
  year = 2020,
  month = oct,
  pages = {1--10},
  issn = {2332-5615},
  doi = {10.1109/AIPR50011.2020.9425190},
  urldate = {2025-09-28},
  langid = {american}
}

@inproceedings{dongEvadingDefensesTransferable2019,
  title = {Evading {{Defenses}} to {{Transferable Adversarial Examples}} by {{Translation-Invariant Attacks}}},
  booktitle = {Proceedings of the {{IEEE}}/{{CVF Conference}} on {{Computer Vision}} and {{Pattern Recognition}}},
  author = {Dong, Yinpeng and Pang, Tianyu and Su, Hang and Zhu, Jun},
  year = 2019,
  pages = {4312--4321},
  urldate = {2026-01-31}
}

@inproceedings{linNesterovAcceleratedGradient2019,
  title = {Nesterov {{Accelerated Gradient}} and {{Scale Invariance}} for {{Adversarial Attacks}}},
  booktitle = {International {{Conference}} on {{Learning Representations}}},
  author = {Lin, Jiadong and Song, Chuanbiao and He, Kun and Wang, Liwei and Hopcroft, John E.},
  year = 2019,
  month = sep,
  urldate = {2026-01-31},
  langid = {english}
}

@inproceedings{liuDelvingTransferableAdversarial2017,
  title = {Delving into {{Transferable Adversarial Examples}} and {{Black-box Attacks}}},
  booktitle = {International {{Conference}} on {{Learning Representations}}},
  author = {Liu, Yanpei and Chen, Xinyun and Liu, Chang and Song, Dawn},
  year = 2017,
  month = feb,
  urldate = {2026-01-31},
  langid = {english}
}

@inproceedings{wangEnhancingTransferabilityAdversarial2021,
  title = {Enhancing the {{Transferability}} of {{Adversarial Attacks Through Variance Tuning}}},
  booktitle = {Proceedings of the {{IEEE}}/{{CVF Conference}} on {{Computer Vision}} and {{Pattern Recognition}}},
  author = {Wang, Xiaosen and He, Kun},
  year = 2021,
  pages = {1924--1933},
  urldate = {2026-01-31},
  langid = {english}
}

@inproceedings{xieImprovingTransferabilityAdversarial2019,
  title = {Improving {{Transferability}} of {{Adversarial Examples With Input Diversity}}},
  booktitle = {Proceedings of the {{IEEE}}/{{CVF Conference}} on {{Computer Vision}} and {{Pattern Recognition}}},
  author = {Xie, Cihang and Zhang, Zhishuai and Zhou, Yuyin and Bai, Song and Wang, Jianyu and Ren, Zhou and Yuille, Alan L.},
  year = 2019,
  pages = {2730--2739},
  urldate = {2026-01-31}
}
%%% and comment out the ``thebibliography'' section.

%%% Comment out this section when you \bibliography{references} is enabled.

\end{document}